\documentclass[10pt,twocolumn,letterpaper]{article}

\usepackage{cvpr}      
\usepackage{float}

\usepackage{bm}
\usepackage{yhmath, amsthm}
\usepackage{multirow}
\usepackage{bbding}
\usepackage{makecell}

\definecolor{cvprblue}{rgb}{0.21,0.49,0.74}
\usepackage[pagebackref,breaklinks,colorlinks,allcolors=cvprblue]{hyperref}

\def\paperID{11085} 
\def\confName{CVPR}
\def\confYear{2025}

\title{Mimic In-Context Learning for Multimodal Tasks}

\author{Yuchu Jiang\textsuperscript{1,2}  \quad \quad \qquad
\quad Jiale Fu\textsuperscript{1,2}  \quad \quad \quad \quad
\quad Chenduo Hao\textsuperscript{1,2} \quad \quad \quad \quad
\quad Xinting Hu \textsuperscript{3}\qquad  \\
{\tt\small kamichanw@seu.edu.cn 
\quad jiale.fu@seu.edu.cn
\quad 213201447@seu.edu.cn
\quad xinting001@e.ntu.edu.sg
}
\vspace{0.7em}
\\
\quad Yingzhe Peng\textsuperscript{1,2}  \quad \quad \quad 
\quad Xin Geng\textsuperscript{1,2}  \quad \quad \quad \quad
\quad Xu Yang\textsuperscript{1,2}\thanks{Corresponding Author}
\qquad \qquad\\
{\tt\small 
yingzhe.peng@seu.edu.cn
\quad xgeng@seu.edu.cn
\quad xuyang\_palm@seu.edu.cn
}
\vspace{0.7em}\\
\textsuperscript{1}Southeast University \\
\textsuperscript{2}Key Laboratory of New Generation Artificial Intelligence Technology and \\
Its Interdisciplinary Applications, (Southeast University), Ministry of Education\\
\textsuperscript{3}Nanyang Technological University\\
}
\begin{document}
\maketitle
\begin{abstract}

Recently, In-context Learning (ICL) has become a significant inference paradigm in Large Multimodal Models (LMMs), utilizing a few in-context demonstrations (ICDs) to prompt LMMs for new tasks. However, the synergistic effects in multimodal data increase the sensitivity of ICL performance to the configurations of ICDs, stimulating the need for a more stable and general mapping function. Mathematically, in Transformer-based models, ICDs act as ``shift vectors'' added to the hidden states of query tokens. Inspired by this, we introduce \textbf{Mim}ic \textbf{I}n-\textbf{C}ontext Learning (\textbf{MimIC}) to learn stable and generalizable shift effects from ICDs. Specifically, compared with some previous shift vector-based methods, MimIC more strictly approximates the shift effects by integrating lightweight learnable modules into LMMs with four key enhancements: 1) inserting shift vectors after attention layers, 2) assigning a shift vector to each attention head, 3) making shift magnitude query-dependent, and 4) employing a layer-wise alignment loss. Extensive experiments on two LMMs (Idefics-9b and Idefics2-8b-base) across three multimodal tasks (VQAv2, OK-VQA, Captioning) demonstrate that MimIC outperforms existing shift vector-based methods. The code is available at \url{https://github.com/Kamichanw/MimIC}.

\end{abstract}    
\section{Introduction}
In-Context Learning (ICL) allows models to generalize from a few examples, known as in-context demonstrations (ICDs), enabling them to learn new tasks without explicit fine-tuning~\cite{iclsurvey,gpt3,radford2019language,yin2023survey,luo2024context}. This approach has become a significant inference paradigm for both Large Language Models (LLMs) and Large Multimodal Models (LMMs)~\cite{peng2025lmm} and finds broad applications in areas such as recommendation systems~\cite{guo2024scaling,yin2024dataset,shen2024optimizing} and point cloud understanding~\cite{jiang2024dgpic}. However, ICL in LMMs faces more limitations than LLMs due to the synergistic effects of integrating vision and language data~\cite{cfg-icd,yi2024bridge}, making some strategies useful in LLM lose their efficacy. For example, while various studies in LLM show that using similar ICDs as the query is beneficial~\cite{icd-sim-selection,icd-sim-selection2}, ~\cite{wyl} found that in captioning tasks, using less similar images may actually improve performance when only low-quality in-context captions are available. This is because similar images can lead LMMs to copy captions through shortcut inference, rather than generalizing to new examples.

Due to synergistic effects, when implementing ICL, it is hard for LMMs to capture the general mapping from input-output pairs of complex multimodal ICDs as LLMs. Instead,~\cite{cfg-icd} shows that LMMs tend to rely on the distribution of ICDs to narrow the prediction space, \eg, in Visual Question Answering (VQA), a LMM might recognize the ICD answer format and respond an answer in the same format, rather than learning the correct function as in language QA. Consequently, the ICL performance in LMMs is more sensitive than in LLMs to ICD configurations~\cite{order-sensitivity,cali-before-use} and finding optimal ICD configurations in LMM is still an open question~\cite{iclsurvey}. A straightforward approach to mitigate the high sensitivity issue is to use more ICDs to help LMMs recognize stable patterns for improved predictions. However, image inputs require more tokens than text, and increasing the number of ICDs significantly raises computational demands. Moreover, current LMMs, like open-source 9B models, typically support only up to 32-shot ICDs~\cite{idefics1}, rendering this approach impractical at larger scales.

In this way, we may wonder whether it is possible to learn a general mapping function from ICDs and then directly incorporate this function into LMMs to enhance ICL performance. Interestingly, from a mathematical perspective, the role of ICDs can be seen as adding shift vectors to the hidden representations of query tokens in LLMs/LMMs~\cite{icv,licv,functionvector,taskvector}. Motivated by this, researchers propose to find a general shift vector as a general mapping function to transform the query space into the answer space. The early methods~\cite{functionvector, taskvector, icv} used heuristic-based approaches to generate shift vectors, which were effective in simple NLP tasks but proved insufficient for more complex multimodal tasks. To address this, a recent study, LIVE~\cite{licv}, introduces a training-based method to learn the shift vector from a large supporting set, outperforming these heuristic-based methods.

\begin{figure}[t]
\centering
\includegraphics[width=\linewidth]{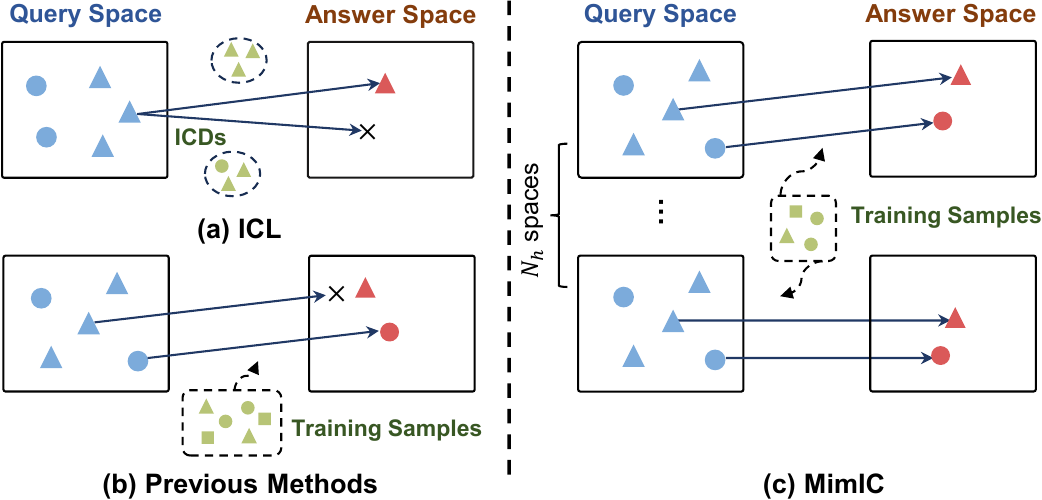}
\caption{Sketches of shift effects from query space to answer space. (a) Traditional ICL induces the shift vector by ICDs, which is sensitive to ICD configurations, \ie, changing one ICD make prediction incorrect. (b) Previous shift vector-based methods insert a query-independent shift vector learned from a large training set, causing equal shift magnitude for diverse queries, which may make prediction incorrect. (c) MimIC assigns a unique query-dependent shift vector learned from fewer training samples after each attention head layer, shifting diverse magnitude for different queries, thus achieving stronger generalization ability.}

\label{fig:shift_vector}
\end{figure}
Although these mathematically inspired methods improve ICL performance, closer examination in \cref{sec:formula} reveals that they use incomplete approximations. 
First, the formula suggests that the shift vector should be applied after the attention layers, but current methods incorrectly place it after the feed-forward network (FFN) layers.
This misuse leads to a second issue: since Transformers use multi-head attention, where each head may have a distinct representation space, applying the shift vector after the FFN overlooks the need for separate shifts for each attention head, reducing the effectiveness of the mapping. Third, the formula implies that the shift magnitude should depend on the query, but these methods use a query-independent shift magnitude. Consequently, as shown in \cref{fig:shift_vector} (b), during inference, the fixed shift magnitude can lead to poor predictions for diverse queries.

In this work, we propose to approximate the shift effect more rigorously to better \textbf{Mim}ic \textbf{I}n-\textbf{C}ontext Learning (\textbf{MimIC}), offering four key improvements for better adaptability and efficiency. First, we position the shift vector after the attention layers instead of the FFN layers. This change allows each attention head to learn a unique shift vector, capturing distinct representation shifts as illustrated in \cref{fig:shift_vector} (c), which is our second enhancement. Third, we make the scaling factor of the shift query-dependent, enabling dynamic adjustment of the shift magnitude during inference.
Finally, we implement a layer-wise alignment loss to ensure that hidden states in zero-shot setting closely align with those of ICL, which allows our method to achieve ICL-like performance with minimal training data.

We validate the effectiveness of MimIC on three foundational multimodal tasks: VQAv2, OK-VQA and Image Captioning (IC) and on two prominent open-source LMMs: Idefics1~\cite{idefics1} and Idefics2~\cite{idefics2}, which represent cross-attention and fully autoregressive architectures, respectively. The results show MimIC surpasses standard ICL, \eg, it achieves a 3.46\% accuracy/3.57\% accuracy/9.00 CIDEr improvement than 32-shot ICL on VQAv2/OKVQA/IC on Idefics1. Moreover, MimIC's generalization allows it to match 32-shot ICL performance with only 1-shot ICL guidance. Compared to previous methods, MimIC achieves superior results, \eg, it improves 4.04\% accuracy /4.99\% accuracy/2.13 CIDEr than the second-best method on VQAv2/OKVQA/IC. Furthermore, comprehensive ablation studies and analyzes confirm the effectiveness of our four proposed enhancements, showing that MimIC requires fewer training samples and achieves a better approximation to ICL compared to other trainable methods.

In summary, we have the following contributions:
\begin{itemize}[leftmargin=.1in]
\item Mathematically, we show the flaws of the approximations in previous shift vector-based methods and propose a more rigorous approximation, offering a stronger mathematical assurance for implementation.
\item Guided by the mathematical formula, we propose a feasible method that achieves approximation by adding fewer learnable parameters.
\item The results of the experiment show that MimIC achieves consistent improvements compared to the original ICL, previous shift vector-based methods, and LoRA in three multimodal tasks on two LMMs.
\end{itemize}

\section{Related Work}
\subsection{In-Context Learning.}
In-Context learning (ICL) refers to a model's ability to perform new tasks by conditioning on a sequence of input-output examples without requiring updates to its parameters~\cite{iclsurvey,gpt3}. This mechanism, widely adopted in Large Language Models (LLMs), allows models to enhance downstream task performance with minimal labeled data~\cite{ft-vs-icl, cot-reasoning}. However, practical applications of ICL face two prominent challenges. First, its performance is highly sensitive to the selection~\cite{cfg-icd,cali-before-use,gao-etal-2021-making,rubin-etal-2022-learning,li2023unified,ye2023compositional} and ordering~\cite{order-sensitivity, wu-etal-2023-self, kumar2021reordering} of in-context demonstrations (ICDs). Methods that select and utilize high-quality ICDs, such as similarity-based retrieval~\cite{icd-sim-selection,rices,icd-sim-selection2,he2023geometric,cao2022image}, are often computationally expensive and not scalable in data-scarce scenarios. Second, an excessive number of demonstrations can result in long context windows, which significantly slows down inference~\cite{long-ctx-cost-time}.

\subsection{ICL in Large Multimodal Models.}
In the realm of multimodal models, several approaches have incorporated ICL capabilities by training on interleaved image-text datasets~\cite{flamingo, idefics1,idefics2, li2024llava, li2024llava-one,parrot}. However, integrating ICL into multimodal models introduces unique challenges that are often underexplored~\cite{zhao2024multi}. Leveraging the inference strengths of LLMs, Large Multimodal Models (LMMs) such as Idefics~\cite{idefics1} and Idefics2~\cite{idefics2} exhibit ICL capabilities by using multiple samples as contextual information during training. Nonetheless, the inherent complexity and diversity of multimodal tasks exacerbate existing challenges in ICL, making it more difficult to fully harness the potential of multimodal ICL~\cite{wyl,cfg-icd,xu2024introspection, baldassini2024makes, luo2024does}.

A recent study by~\cite{lever-lm} shows that for some LMMs, similarity-based retrieval methods for selecting ICDs can perform worse than random selection. This highlights the difficulty of identifying high-quality ICDs in multimodal tasks, which remains an open problem. Moreover, current mainstream multimodal architectures, such as those used in~\cite{llava,idefics2,wang2024qwen2}, typically concatenate image tokens directly with text tokens. Since one image can be equivalently encoded to thousands of text tokens, incorporating multiple image tokens significantly increases the context length, leading to substantial slowdowns in ICL inference.

\subsection{Understanding ICL Mechanisms.}
Understanding the underlying mechanisms of ICL is critical for improving its effectiveness in guiding the inference processes of LLMs. Various approaches have been proposed to explain why ICL works. For instance, \cite{rethink-icd-role} suggests that ICL performance depends not only on the accuracy of true labels but also on factors such as label space representation, input distribution, and sequence format. Additionally, \cite{icl-grad-descent} argues that Transformer attention mechanisms in ICL operate similarly to gradient descent, framing the process as implicit fine-tuning. Another perspective comes from the concept of a ``task vector''~\cite{taskvector}, which posits that ICL compresses training data into a single vector that guides the model’s outputs. Similarly, the ``Function Vector''~\cite{functionvector} idea identifies a compact neural encoding of input-output functions within autoregressive language models.

A prevailing approach interprets ICL through the framework of shift vector, where ICL is understood as encoding task context within a learnable vector representation that modulates model behavior. For example, LIVE~\cite{licv} employs a self-distillation strategy to optimize a learnable ICV directly, which enhances ICL’s performance in multimodal settings. Similarly, the Multimodal Task Vector (MTV) method~\cite{mtv} averages activations across multiple attention heads, encoding task-specific information as task vectors to enable robust few-shot multimodal ICL.
Our work aligns with this methodology but introduces a unique and innovative approach. Unlike previous methods, MimIC achieves superior data efficiency and enhanced learning capability.

\section{Mimicking In-Context Learning}
Our objective is to mimic in-context learning (ICL) from the perspective of space shift induced by ICL with fewer trainable parameters and training samples. We begin by analyzing the behavior of in-context demonstrations (ICDs) within the self-attention mechanism in \cref{sec:formula}. This analysis reveals that the output of self-attention can be decomposed into two components: one affected by the ICDs and the other independent of them. We then detail how to approximate the component influenced by the ICDs to capture the general shift effect in \cref{sec:mimic}.

\subsection{Mathematic Analyses}
\label{sec:formula}
ICL allows large language models (LLMs) or large multimodal models (LMMs) to generalize to new tasks by providing a few ICDs directly in the input. Formally, the prompt context is defined as $\bm{C} = \{\bm{X}_D, \bm{X}\}$, where $\bm{X}_D = \{\bm{X}_1, \bm{X}_2, \dots, \bm{X}_k\} \in \mathbb{R}^{l_D \times d}$ represents the concatenation of $k$ ICDs, and $\bm{X} \in \mathbb{R}^{l_q \times d}$ is the query input. Here, $l_D$ and $l_q$ denote the number of tokens in $\bm{X}_D$ and $\bm{X}$, respectively, and $d$ is the embedding dimension.

Multi-head self-attention applies the self-attention (SA) mechanism over $N_h$ heads, each parameterized by weight matrices $\bm{W}_k, \bm{W}_q, \bm{W}_v \in \mathbb{R}^{d \times d_h}$ to project $\bm{C}$ into keys $\bm{K}_C$, queries $\bm{Q}_C$, and values $\bm{V}_C$. Typically, $d_h$ is set to $d/N_h$ to reduce parameter usage by operating each attention head in a lower-dimensional space. For a specific head, the key mapping is defined as:
\begin{equation}
    \bm{K}_C = \bm{C} \bm{W}_k = \begin{bmatrix} \bm{X}_D \\ \bm{X} \end{bmatrix} \bm{W}_k = \begin{bmatrix} \bm{K}_D \\ \bm{K} \end{bmatrix}.
\end{equation}
Similarly, we compute the corresponding $\bm{Q}_D, \bm{Q}$, and $\bm{V}_D, \bm{V}$ using $\bm{W}_q$ and $\bm{W}_v$, respectively.
For each query vector $\bm{q} \in \bm{Q}$, the single-head self-attention operation is\footnote{For simplicity, we illustrate the method using a single head, though each head typically has distinct weight matrices.} :
\begin{align}
\small
&\textrm{SA}\left(\bm{q}, \begin{bmatrix} \bm{K}_D \\ \bm{K} \end{bmatrix}, \begin{bmatrix} \bm{V}_D \\ \bm{V} \end{bmatrix}\right) \notag \\
&= \textrm{softmax}\left( \begin{bmatrix} \bm{q} \bm{K}_D^\top, \bm{q} \bm{K}^\top \end{bmatrix} \right) \begin{bmatrix} \bm{V}_D \\ \bm{V} \end{bmatrix} \notag \\
& = \left[ \frac{\exp\left(\bm{qK}_D^\top\right)}{Z_1+Z_2}, \frac{\exp\left(\bm{qK}^\top\right)}{Z_1+Z_2}\right] \begin{bmatrix} \bm{V}_D \\ \bm{V} \end{bmatrix} \notag \\
&= \frac{Z_2}{Z_1+Z_2} \frac{\exp(\bm{q}\bm{K}^\top)}{Z_2}\bm{V} + \frac{Z_1}{Z_1+Z_2}\frac{\exp(\bm{q}\bm{K}_D^\top)}{Z_1}\bm{V}_D\notag \\
&= \frac{Z_2}{Z_1+Z_2} \textrm{softmax}(\bm{q}\bm{K}^\top)\bm{V} + \frac{Z_1}{Z_1+Z_2}\textrm{softmax}(\bm{q}\bm{K}_D^\top)\bm{V}_D\notag \\
&= (1-\mu) \textrm{SA}(\bm{q}, \bm{K}, \bm{V}) + \mu \textrm{SA}(\bm{q}, \bm{K}_D, \bm{V}_D) \notag \\
&= \underbrace{\textrm{SA}(\bm{q}, \bm{K}, \bm{V})}_{\textrm{standard attention}} + \underbrace{\mu (\textrm{SA}(\bm{q}, \bm{K}_D, \bm{V}_D))-\textrm{SA}(\bm{q}, \bm{K}, \bm{V})}_{\textrm{shift vector}}
\label{eq:expand-head}
\end{align}
where $\mu$ is a scalar representing the normalized attention weights over the ICDs:
\begin{align}
    \mu(\bm{q}, \bm{K}_D, \bm{K})=\frac{Z_1(\bm{q},\bm{K}_D)}{Z_1(\bm{q},\bm{K}_D) + Z_2(\bm{q},\bm{K})}, 
\label{eq:mu}
\end{align}
where $Z_1(\bm{q},\bm{K}_D) = \sum_{i=1}^{l_D} \exp(\bm{q} \bm{K}_D^\top)_i$ and $Z_2(\bm{q},\bm{K}) = \sum_{j=1}^{l_q} \exp(\bm{q} \bm{K}^\top)_j$.

\cref{eq:expand-head} shows that the self-attention over the prompt context $\bm{C}$ can be decomposed into two terms. For the former ``standard attention'', it is the self-attention over the query tokens, which is independent of the ICDs.  While for the latter ``shift vector'', it is the shift effects caused by the ICDs to shift the query space into the answer space, and such effects is calculated as the attention between the ICDs and the query $\bm{q}$. This shift is governed by the attention difference term $\textrm{SA}(\bm{q}, \bm{K}_D, \bm{V}_D) - \textrm{SA}(\bm{q}, \bm{K}, \bm{V})$ and the scalar $\mu(\bm{q}, \bm{K}_D, \bm{K})$, both of which depend on the ICDs. 

Now, we show which terms in \cref{eq:expand-head} are affected by ICDs and the fluctuation of ICDs changes these terms, making the predictions sensitive to the ICD configurations. Moreover, directly applying self-attention in LLMs or LMMs over long ICD inputs costs substantial computation burdens. In the next section, we describe how to approximate the ICD-affected terms in \cref{eq:expand-head} to capture the general shift effect, improving robustness in ICL and significantly increasing inference efficiency.
\begin{figure}[t]
  \centering
   \includegraphics[width=0.45 \textwidth]{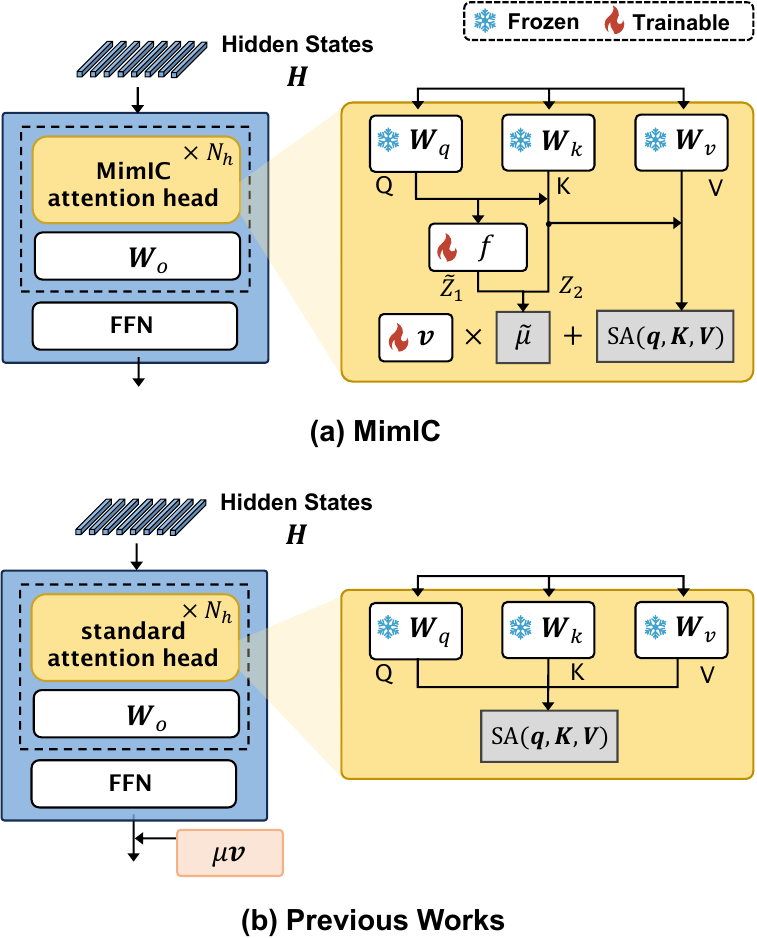}
   \caption{Comparison of MimIC and previous shift vector based methods. (a) MimIC changes the attention mechanism for each head, which inserts a learnable shift vector $\bm{v}$ with a query-dependent magnitude $\mu$. (b) Previous methods insert the pre-calculated or learnable shift vector with a query-independent $\mu$ after FFN layer without changing the attention mechanism. }

   \label{fig:mimic-attn}
\end{figure}
\subsection{Mimicking ICD Affected Terms}
\label{sec:mimic}
From ~\cref{eq:expand-head} and ~\cref{eq:mu}, we observe that only $\textrm{SA}(\bm{q}, \bm{K}_D, \bm{V}_D)$ and $Z_1(\bm{q}, \bm{K}_D)$ are affected by ICDs. To mimic ICL, we approximate these terms by inserting a few lightweight modules into the attention heads of LMM and name them as \textbf{MimIC Attention Heads} as in \cref{fig:mimic-attn}(a). 

To approximate $Z_1(\bm{q}, \bm{K}_D)$, we note that it is a positive scalar dependent solely on the current query token $\bm{q}$ and the ICD keys $\bm{K}_D$. Therefore, we use a simple mapping: a trainable linear layer $f(\cdot): \mathbb{R}^{d_h} \to \mathbb{R}$ to approximate $\log Z_1$. For the attention difference term $\textrm{SA}(\bm{q}, \bm{K}_D, \bm{V}_D) - \textrm{SA}(\bm{q}, \bm{K}, \bm{V})$, as \cref{fig:mimic-attn}(a) shows, we insert a learnable vector $\bm{v} \in \mathbb{R}^{d_h}$ in each attention head to capture the general shift effect for this head. Then, the output of~\cref{eq:expand-head} in MimIC attention head is computed as $
\operatorname{SA}(\bm{q}, \bm{K}, \bm{V}) + \tilde{\mu}(\bm{q}, \bm{K}) \bm{v}$,
where  $\tilde{\mu}(\bm{q}, \bm{K}) = \tilde{Z_1}(\bm{q})/(\tilde{Z_1}(\bm{q})+Z_2(\bm{q}, \bm{K}))$ and $\tilde{Z_1}(\bm{q}) = \exp(f(\bm{q}))$.
After obtaining the outputs from all MimIC attention heads, they are concatenated, flattened, and passed through the matrix $\bm{W}_o \in \mathbb{R}^{d \times d}$ and FFN layer.

\begin{figure}[t]
  \centering
   \includegraphics[width=\linewidth]{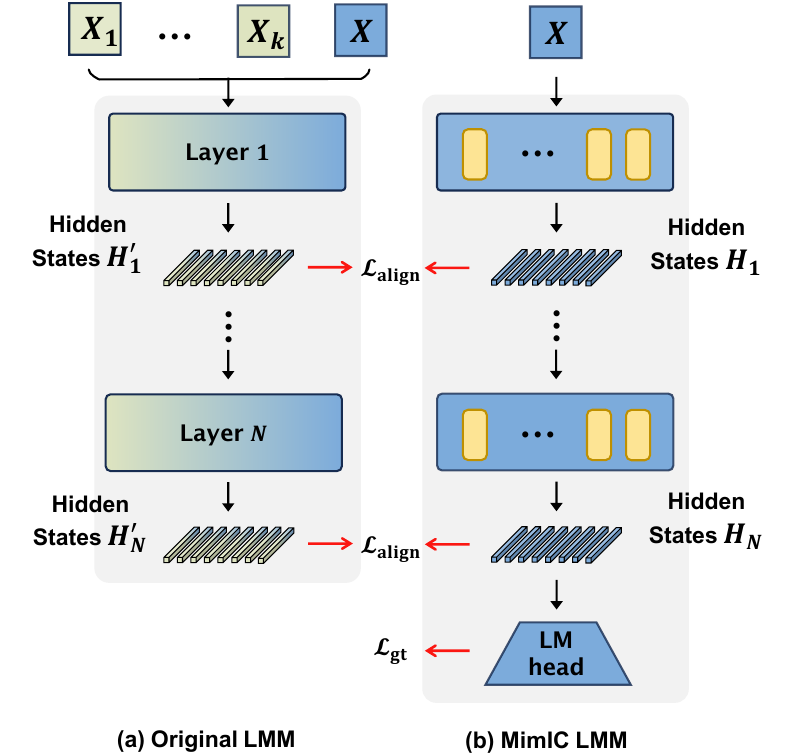}
   \caption{Overall training framework of MimIC. (a) The original LMM processes $k$ ICDs and query input as conventional ICL, generating hidden states $\bm{H}_1^\prime$ to $\bm{H}_N^\prime$ at each layer. (b) In MimIC LMM, only a single query input $\bm{X}$ is processed, producing shifted hidden states $\bm{H}_1$ to $\bm{H}_N$, which are aligned with the original hidden states via the alignment loss $\mathcal{L}_\text{align}$. Additionally, the logits of language modeling head is used to compute ground truth loss $\mathcal{L}_\text{gt}$. The yellow blocks represents MimIC attention heads.}
   \label{fig:framework}
\end{figure}

Given this MimIC attention head, we replaces all the self-attention heads of the original LMM to get the \textbf{MimIC LMM} as in \cref{fig:framework}(b). Then we hope MimIC LLM can handle a single query $\bm{X}$ in the same way the original LMM implements ICL, \ie, using the ICDs $\bm{X}_D$ to produce the result for $\bm{X}$. To achieve this, given a training set, we randomly select $k$ samples as ICDs $\bm{X}_D$ and one sample as the query $\bm{X}$. As \cref{fig:framework}(a) shows, for the original LMM, we input the context $\bm{C} = \{\bm{X}_D, \bm{X}\}$ into it to get the hidden states at each layer, which are recorded as $\bm{\mathcal{H}}^\prime = \{\bm{H}_1^\prime, \dots, \bm{H}_N^\prime\}$. For MimIC LMM, we only input $\bm{X}$ into it to get the hidden states $\bm{\mathcal{H}} = \{\bm{H}_1, \dots, \bm{H}_N\}$. To make MimIC LMM behave similar as the original LMM, we set an alignment loss $\mathcal{L}_{\textrm{align}}$ to make $\bm{\mathcal{H}}$ be close to $\bm{\mathcal{H}}^\prime$. Specifically, $\mathcal{L}_{\textrm{align}}$ is computed as the average $L_2$ distance between the hidden states at each layer, ensuring a layer-wise contribution:
\begin{equation}
\mathcal{L}_\text{align} = \frac{1}{N}\sum_{i=1}^{N} \sum_{j=1}^{l_q} \left\lVert \bm{h}_{i,j} - \bm{h}_{i,j}^\prime \right\rVert_2^2.
\end{equation}

In addition, we employ the language modeling loss $\mathcal{L}_{\textrm{gt}}$ to enhance the model's performance on downstream tasks, allowing it to learn task-specific features more effectively. Thus, the total loss function is:
\begin{equation}
    \mathcal{L} = \mathcal{L}_\text{align} + \lambda \mathcal{L}_\text{gt},
    \label{eq:loss}
\end{equation}
where $\lambda$ is a hyperparameter that controls the trade-off between alignment and task-specific loss.

During training, since the ICDs are randomly selected in each step, MimIC LMM is encouraged to capture the most general shift pattern from the fluctuated shifts brought by various random configurations of ICDs. After training, the attention difference term $\textrm{SA}(\bm{q}, \bm{K}_D, \bm{V}_D) - \textrm{SA}(\bm{q}, \bm{K}, \bm{V})$ captures the general shift direction across various ICD configurations, while $Z_1(\bm{q}, \bm{K}_D)$ adjusts the shift magnitude based on the query input $\bm{q}$. Consequently, when using MimIC to inference, ICDs are no longer required, leading to a significant improvement in inference speed.

\subsection{Design Difference from Previous Methods}
~\cref{fig:mimic-attn} compares the differences between MimIC and previous shift vector-based methods. First, previous methods insert the shift vector $\bm{v}$  after FFN layer, while we insert $\bm{v}$ into each attention head. In this way, each vector can learn suitable shift direction for the corresponding head representation space, leading to more powerful shift effects. Second, previous methods use query-independent shift magnitude $\mu$, while we set $\mu$ be depended on the query, enabling dynamic adjustment of the shift magnitude to achieve better performance.
Although these differences seem subtle, experiments will show that the devil is in the details. The findings presented in ~\cref{sec:ablation} highlight that MimIC's multi-head, query-dependent magnitude is essential for capturing general shift effects from distinct representation space.

\section{Experiment}
\subsection{Setup}
\noindent\textbf{Models, datasets, and metrics.} 
We evaluate MimIC on two large-scale multimodal models (LMMs), Idefics-9b~\cite{idefics1} and Idefics2-8b-base~\cite{idefics2}, referred to as Idefics1 and Idefics2, across three datasets: VQAv2~\cite{vqav2}, OK-VQA~\cite{okvqa}, and COCO Caption~\cite{cococap}. Idefics1 is based on a cross-attention architecture, while Idefics2 employs a fully autoregressive architecture. These models represent two popular architectures for vision-language models. For each dataset, we randomly select 1,000 samples for training. We follow the evaluation protocol of previous works~\cite{licv, cfg-icd}, using 10,000 validation samples from VQAv2 and the full validation splits for OK-VQA and COCO. We present more results on various datasets in Appendix.

\noindent\textbf{Implementation details.} 
During each training step, we randomly select 32 samples as ICDs for Idefics1 and 8 for Idefics2, with one additional distinct sample as the query input. We employ the AdamW optimizer with a learning rate of $5 \times 10^{-3}$, coupled with a cosine annealing scheduler with warmup, allocating 10\% of the total steps for warmup. The value of $\lambda$ in \cref{eq:loss} is set to 0.5. All results are reported from the best-performing epoch. Additional implementation details are provided in the Appendix.

\subsection{Comparison with Existing Methods}
\label{sec:main-result}
\begin{table}[t]
\centering
\resizebox{\linewidth}{!}{
\begin{tabular}{@{}cccccc@{}}
\toprule
\textbf{Model} & \textbf{Method} & \textbf{\# Params (M)} & \textbf{VQAv2} & \textbf{OK-VQA} & \textbf{COCO} \\ \midrule
\multirow{8}{*}{\rotatebox{90}{Idefics-9b}} & Zero-shot & - & 29.25 & 30.54 & 63.06 \\
 & 32-shot ICL & - & 56.18 & 48.48 & 105.89 \\
 & RICES & - & 58.07 & 51.11 & 110.64 \\ \cmidrule[0.1pt]{2-6} 
 & FV & - & 30.21 & 31.02 & 74.01 \\
 & TV & - & 43.68 & 32.68 & 84.72 \\ \cmidrule[0.1pt]{2-6} 
 & LIVE & 0.13 ($\times 0.5$) & 53.71 & 46.05 & \underline{112.76} \\
 & LoRA & 25.0 ($\times 96.2$) & \underline{55.60} & \underline{47.06} & 97.75 \\
 & MimIC & 0.26 ($\times 1.0$) & \textbf{59.64} & \textbf{52.05} & \textbf{114.89} \\ \midrule
\multirow{8}{*}{\rotatebox{90}{Idefics2-8b-base}} & Zero-shot & - & 55.39 & 43.08 & 40.00 \\
 & 8-shot ICL & - & 66.20 & 57.68 & 122.51 \\
 & RICES & - & 66.44 & 55.73 & 111.44 \\ \cmidrule[0.1pt]{2-6} 
 & FV & - & 36.47 & 34.58 & 75.24 \\
 & TV & - & 47.12 & 38.27 & 87.61 \\ \cmidrule[0.1pt]{2-6} 
 & LIVE & 0.13 ($\times 0.5$) & \underline{67.60} & 54.86 & \underline{126.04} \\
 & LoRA & 17.6 ($\times 67.7$) & 66.54 & \underline{55.05} & 116.69 \\
 & MimIC & 0.26 ($\times 1.0$) & \textbf{69.29} & \textbf{58.74} & \textbf{132.87} \\ \bottomrule
\end{tabular}
}
\caption{
Results of VQAv2, OK-VQA, and COCO on Idefics-9b and Idefics2-8b-base. \textbf{Bold numbers}/\underline{underlined numbers} represent the best/second-best results, respectively.
}

\label{tab:main-result}
\end{table}
\begin{figure*}[ht]
    \centering
    \includegraphics[width=\linewidth]{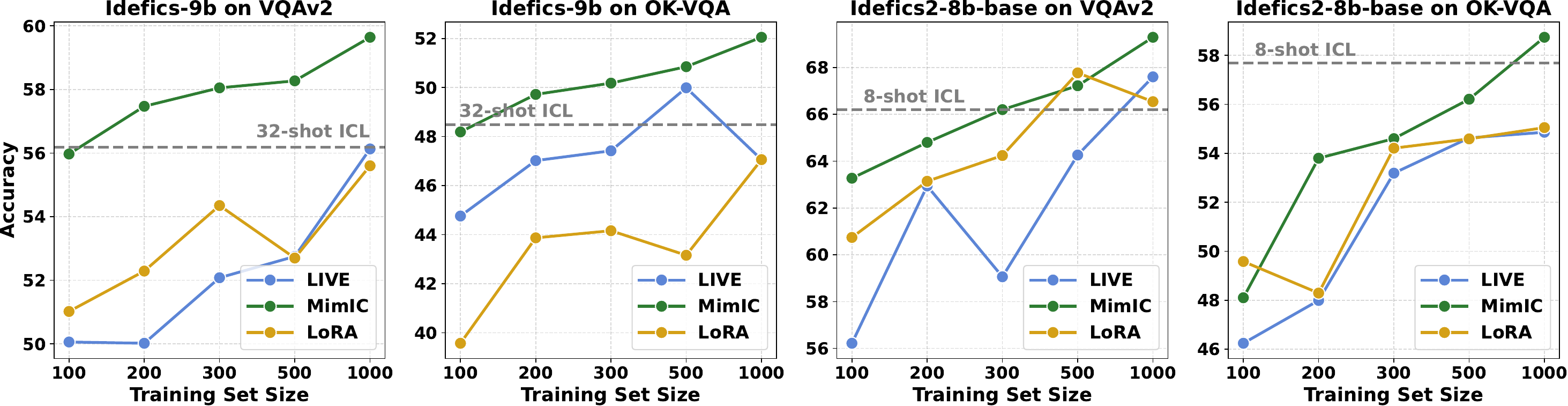}
    \caption{Performance comparisons of trainable methods on two LMMs across VQAv2/OK-VQA with fewer training set size.}
\label{tab:fewer-samples}
\end{figure*}

\noindent\textbf{Compared methods.} We compare MimIC with the following methods: 
(1) \textbf{In-Context Learning (ICL)} is evaluated under three settings: zero-shot, few-shot, and Retrieval-based In-Context Examples Selection (RICES). For few-shot ICL, we use 32/8-shot for Idefics1 and Idefics2, respectively\footnote{We implement up to 8-shot ICL on Idefics2, as it requires more image tokens compared to Idefics1, which far exceeds our computational capacity.}. RICES~\cite{rices} retrieves similar images from the support set for each query image by comparing visual features extracted from a frozen pretrained visual encoder.
(2) \textbf{Task Vector (TV)}~\cite{taskvector} and \textbf{Function Vector (FV)}~\cite{functionvector} extract compact vectors from a set of demonstrations, which are added to the hidden states of the last token in one or more layers. We evaluate these methods across different layers and report the configuration yielding the best performance.
(3) \textbf{Learnable In-Context Vector (LIVE)}~\cite{licv} introduces learnable vectors after each FFN layer, trained with a pipeline similar to MimIC's under the same few-shot setting.
(4) \textbf{LoRA}~\cite{lora} fine-tunes the model by adding low-rank adapters to the attention weights. We apply the widely used configuration, modifying $\bm{W}_q$, $\bm{W}_k$, $\bm{W}_v$, and $\bm{W}_o$ in all attention layers of both vision and language models. 
All trainable methods are trained using 1000 samples. All methods are evaluated using the optimal hyper-parameters recommended in their respective original works, ensuring a fair comparison.

\noindent\textbf{Results analysis.}
\cref{tab:main-result} presents the results of MimIC compared to various baselines across two LMMs and three datasets. In ICL, the performance of RICES significantly differs from the random selection of ICDs, indicating that the choice of ICDs has a substantial impact on ICL performance. Although RICES outperforms random selection across all three datasets on Idefics1, its performance on Idefics2, OK-VQA, and COCO is inferior to that of the random selection method. This suggests that effective ICD selection strategies differ across models.

For non-trainable methods, while they outperform zero-shot baselines on Idefics1, there is still a significant gap compared to 32-shot ICL performance. On Idefics2, these non-trainable methods fail to surpass the zero-shot baseline. This indicates that non-trainable methods are not only ineffective at capturing essential task-specific information but also perform poorly across different LMMs.

Trainable methods consistently improve performance across both LMMs, approaching the effectiveness of few-shot ICL. LIVE performances similar to LoRA with fewer parameters, while its reliance on a fixed shift magnitude during inference limits its ability to generalize across different queries, making it less effective than MimIC. However, for MimIC, on Idefics1, the performance on VQA/captioning improved by an average of 3.52/9.00 compared to 32-shot ICL, and by 4.52/2.13 compared to the second-best method. On Idefics2, MimIC was the only method to consistently outperform ICL, with average improvements of 1.31/10.36 on the VQA and captioning, respectively. Such comparisons validate the powerful of MimIC in improving the ICL performance. Also, MimIC achieves the best results on both LMMs with diverse architectures, indicating greater stability than other methods.

\noindent\textbf{Training with fewer samples.} We conduct further evaluations of trainable methods using a reduced number of training samples. As illustrated in \cref{tab:fewer-samples}, on Idefics1, MimIC only requires 200 samples to exceed the performance of 32-shot ICL. In contrast, other trainable methods typically require a larger number of samples to achieve comparable ICL effectiveness. These comparisons suggest that using a more precise approximation of ~\cref{eq:expand-head} requires fewer training samples to capture the general shift effect of ICL. For instance, compared to LIVE, we were able to exceed its highest performance using only about 1/8 of the data it required for training~\cite{licv}.
\subsection{Ablations and More Analyses}
\label{sec:ablation}
We analyze the effects of various settings on Idefics1, including the necessity of employing a multi-head, query-dependent shift magnitude $\mu$, and the impact of the diverse ICD shot numbers used as ICL guidance during training. Also, we compare the alignment distances between various methods and 32-shot ICL and the hallucinations generated by diverse methods.

\begin{figure*}[th]
    \centering
    \includegraphics[width=0.3\linewidth]{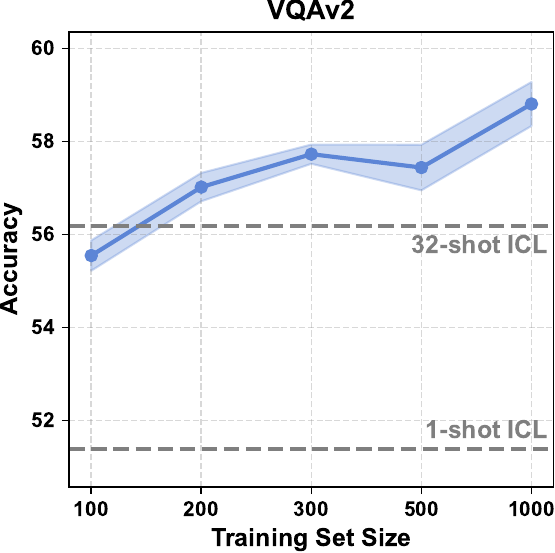}
    \includegraphics[width=0.3\linewidth]{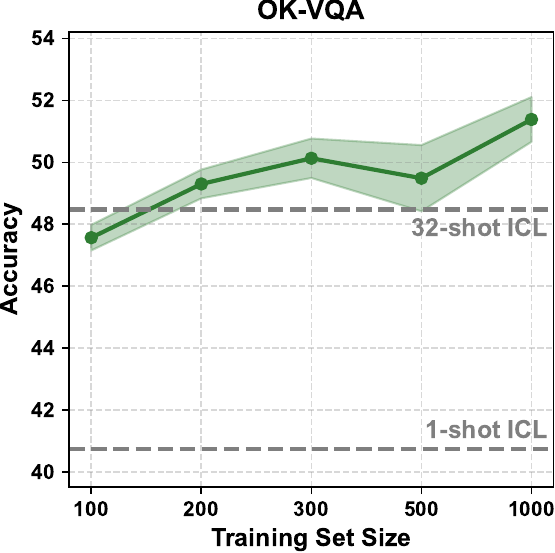}
    \includegraphics[width=0.3\linewidth]{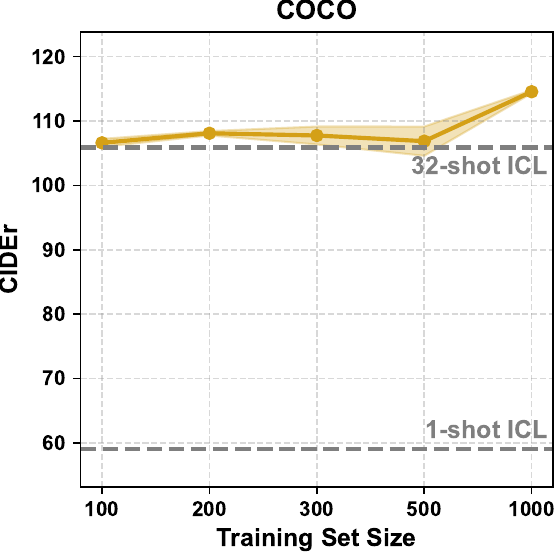}
    \caption{Performance of MimIC trained with varying ICD shots on Idefics-9b, with the shaded area indicating the standard deviation across 1, 4, 8, 16 and 32 shot settings.}
    \label{fig:diff-shot}
\end{figure*}

\noindent\textbf{Effect of multi-head and query-dependent shift magnitude \(\mu\).} We compare two additional settings: (1) \textbf{Head-sharing \(\mu\)}: This involves replacing the original linear layer \(f: \mathbb{R}^{d_h} \rightarrow \mathbb{R}\) of each head with a new linear layer \(g: \mathbb{R}^{d} \rightarrow \mathbb{R}\) that aggregates the information from \(N_h\) heads. As a result, all heads use the query-dependent shift magnitude \(\mu\) produced by this linear layer. (2) \textbf{Query-sharing \(\mu\)}: In this setting, the function \(f\) is removed for each head, and a learnable coefficient \(\mu\) is introduced, leading to a fixed shift magnitude for each query token.

The results in \cref{tab:ablation-mu} show that MimIC outperforms both head-sharing \(\mu\) and query-sharing \(\mu\), suggesting that the multi-head and query-dependent shift magnitude not only capture features from different representation spaces but also that this dynamic behavior, which varies depending on the query, enhances generalization across diverse inputs.

\begin{table}[t]
\centering
\begin{tabular}{@{}cccc@{}}
\toprule
\multicolumn{1}{c}{\textbf{Method}} & \textbf{VQAv2} & \textbf{OK-VQA} & \textbf{COCO} \\ \midrule
Head-sharing $\mu$ & 57.89 & 50.86 & 111.98 \\
Query-sharing $\mu$ & 57.95 & 50.94 & 112.48 \\
MimIC & \textbf{59.64} & \textbf{52.05} & \textbf{114.89} \\ \bottomrule
\end{tabular}
\caption{Performance comparisons with different settings.}
\label{tab:ablation-mu}
\end{table}

\noindent\textbf{Number of ICD shots.} We examine the effect of varying the number of ICD shots on MimIC's performance during training. As shown in \cref{fig:diff-shot}, there is a significant performance gap between 1-shot and 32-shot ICL. ICL performance is highly dependent on the number of demonstrations; when demonstrations are insufficient, the LMMs may misalign query representations. Although MimIC employs ICL as a guiding mechanism, its performance remains largely unaffected by different ICD configurations, demonstrating stability across varying training set sizes. This suggests that MimIC is able to learn the general shift of query representations from the demonstrations, thereby mitigating the negative impact of insufficient demonstrations during training and resulting in a more robust model capable of effectively extracting key task information.

\begin{figure*}[t]
    \centering
    \includegraphics[width=0.9\linewidth]{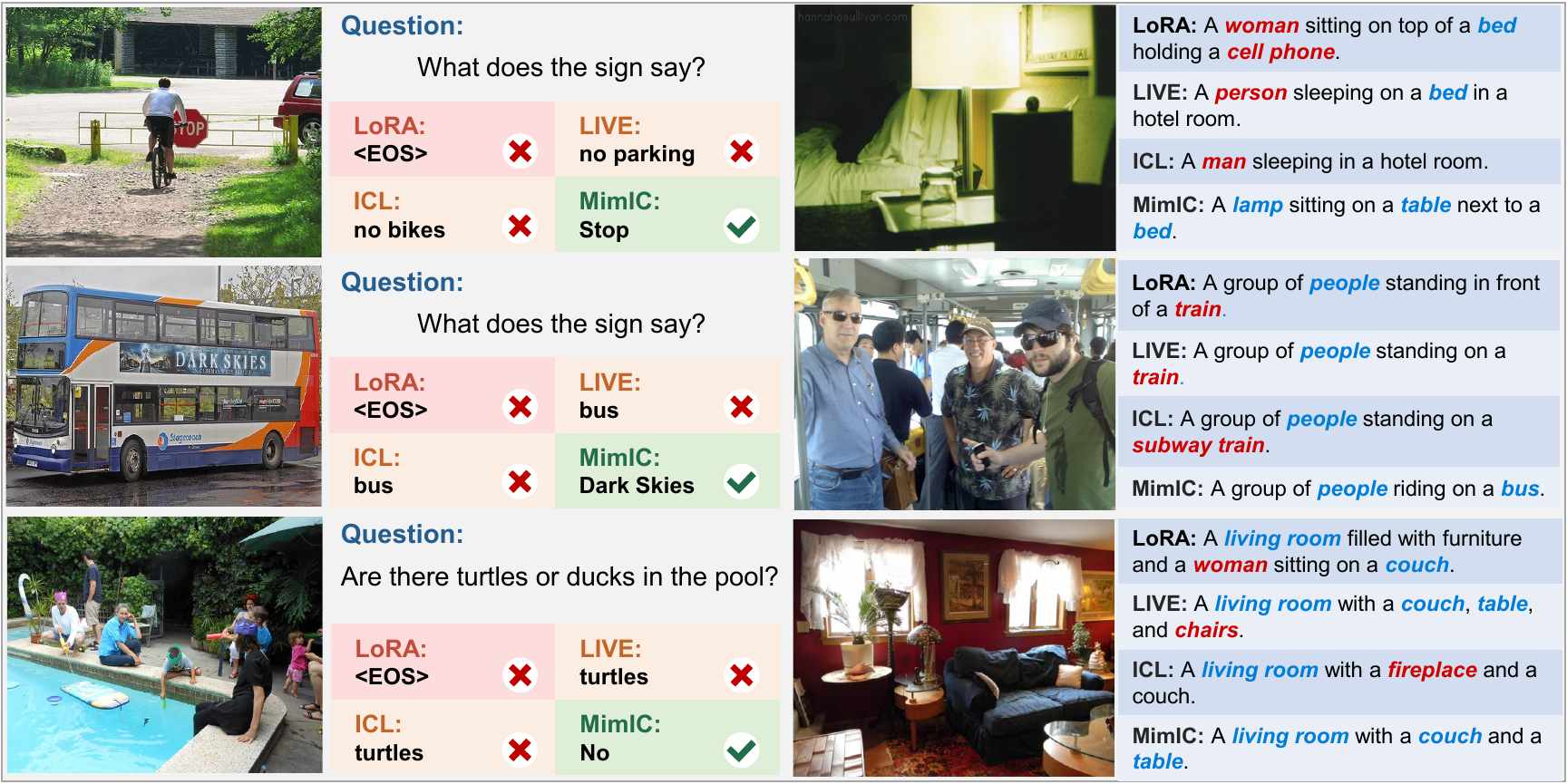}
    \caption{The visualizations of the cases where other methods appear hallucinations on visual question answering (left) and image captioning task (right). The \textcolor[rgb]{0.81,0,0}{\textbf{\textit{red}}} and \textcolor[rgb]{0,0.49,0.85}{\textbf{\textit{blue}}} words represent hallucination objects and correct objects, respectively.}
    \label{fig:case-study-vqa}
\end{figure*}

\noindent\textbf{Alignment effect in latent space.}
\label{sec:align}
\begin{table}[t]
\centering
\resizebox{0.9\linewidth}{!}{
\begin{tabular}{@{}ccccc@{}}
\toprule
 & \textbf{Zero-shot} & \textbf{LIVE} & \textbf{MimIC}$^\dagger$ & \textbf{MimIC} \\ \midrule
VQAv2 & 42.97 & 33.79 & 32.13 & \textbf{30.17} \\
OK-VQA & 41.21 & 34.12 & 29.76 & \textbf{28.25} \\ \bottomrule
\end{tabular}
}
\caption{L2 distance between 32-shot ICL and various methods.}
\label{tab:dist}
\end{table}
Here, we quantitatively assess whether the MimIC attention heads and \(\mathcal{L}_{\textrm{align}}\) proposed in \cref{sec:mimic} facilitate alignment with ICL. Using 200 samples from VQAv2 and OK-VQA, we computed the average L2 distance of the latent representations of the first answer token at each layer, compared to the 32-shot ICL. We also test a variant, MimIC\(^\dagger\), in which \(\mathcal{L}_{\textrm{align}}\) is replaced with KL divergence, as used in LIVE. The results in \cref{tab:dist} demonstrate that MimIC exhibited the smallest distance to 32-shot ICL. Specifically, MimIC is closer to 32-shot ICL than MimIC\(^\dagger\), suggesting that \(\mathcal{L}_{\textrm{align}}\) more effectively enables the shift vectors to capture the characteristics of the ICL shift. Additionally, MimIC\(^\dagger\) showed a smaller distance to the LIVE, highlighting that the mimic attention heads are better able to mimic ICL more precisely.

\begin{table}[t]
\centering
\setlength{\tabcolsep}{3pt}
\renewcommand\arraystretch{1.1}
\resizebox{\linewidth}{!}{
\begin{tabular}{@{}cccccccc@{}} 
\toprule 
 & \textbf{Zero-shot} & \textbf{32-shot ICL} & \textbf{TV} & \textbf{FV} & \textbf{LoRA} & \textbf{LIVE} & \textbf{MimIC} \\ 
\midrule 
\textbf{CHAIRs $\downarrow$} & \textbf{5.93} & 16.78 & 8.88 & 28.26 & 17.42 & 8.65 & 8.51 \\ 
\textbf{CHAIRi $\downarrow$} & \textbf{5.58} & 9.77 & 7.50 & 25.44 & 11.55 & 6.05 & 5.74 \\ 
\textbf{Recall $\uparrow$} & 30.72 & 42.59 & 36.22 & 27.69 & 42.93 & 42.84 & \textbf{43.30} \\ 
\bottomrule 
\end{tabular}
}
\caption{Caption hallucination metrics on various methods.}
\label{tab:cap-hallucination}
\end{table}

\begin{table}[t]
\centering
\setlength{\tabcolsep}{3pt}
\renewcommand\arraystretch{1.1}
\resizebox{\linewidth}{!}{
\begin{tabular}{@{}ccc|cc|cc|cc@{}}
\toprule
 & \multicolumn{2}{c|}{\textbf{4 shot}} & \multicolumn{2}{c|}{\textbf{8 shot}} & \multicolumn{2}{c|}{\textbf{16 shot}} & \multicolumn{2}{c}{\textbf{32 shot}} \\
 & ICL & MimIC & ICL & MimIC & ICL & MimIC & ICL & MimIC \\ \midrule
\textbf{CHAIRs $\downarrow$} & 5.57 & \textbf{4.09} & 5.41 & \textbf{4.55} & 5.56 & \textbf{4.20} & 9.77 & \textbf{5.74} \\
\textbf{CHAIRi $\downarrow$} & 7.20 & \textbf{5.39} & 7.00 & \textbf{6.19} & 7.70 & \textbf{5.52} & 16.78 & \textbf{8.51} \\
\textbf{Recall $\uparrow$} & 39.74 & \textbf{39.84} & 41.0 & \textbf{41.62} & 42.12 & \textbf{42.69} & 42.59 & \textbf{43.30} \\ \bottomrule
\end{tabular}
}
\caption{The hallucination metrics on the image captioning task for MimIC trained with different shot numbers and the corresponding shot number of ICL.}
\label{tab:incr-shot-hallucination}
\end{table}

\noindent\textbf{Hallucinations.}
\cref{fig:case-study-vqa} presents cases where MimIC responds correctly while other methods fail. We also quantitatively analyze hallucinations in image captioning using CHAIRi and CHAIRs\cite{chair}, which measure the proportion of hallucinated words. As shown in ~\cref{tab:cap-hallucination}, MimIC generates fewer hallucinations than non-zero-shot methods while maintaining a high recall rate. Compared to zero-shot, MimIC has a slight increase in hallucinations, which is due to the limitations of ICL, as noted by ~\cite{shukor2023beyond}, where more shots amplify hallucinations. Despite this, MimIC still shows strong hallucination suppression. We also analyze hallucination levels in MimIC with varying shot counts and compare them to ICL models. The results in ~\cref{tab:incr-shot-hallucination} show that hallucinations in MimIC increase with shot count but remain lower than ICL. Notably, MimIC with four shots has a lower hallucination rate than the zero-shot setting and improves recall significantly. This is due to a more precise approximation of the ICL mechanism, outperforming previous shift-based methods.

\section{Conclusion}
Motivated by the insight that in-context demonstrations function as shift vectors applied to the hidden states of query tokens, we propose \textbf{Mim}ic \textbf{I}n-\textbf{C}ontext Learning (\textbf{MimIC}) to learn this effect in Large Multimodal Models (LMMs). MimIC operates by inserting distinct shift vectors into different heads of attention layers, employing a linear layer to generate query-dependent magnitudes for these shift vectors, and leveraging a layer-wise alignment loss to align with in-context learning (ICL). Empirical evaluations across three diverse tasks using two LMMs demonstrate MimIC achieves competitive few-shot ICL performance with significantly reduced inference latency compared to traditional ICL methods, requires fewer training samples relative to LIVE, consistently surpasses other shift vector-based approaches, utilizes fewer parameters than LoRA yet yields superior results, and substantially reduces hallucinations compared to existing techniques.
\section{Acknowledgment}
This work is supported by the National Science Foundation of China (62206048), the Natural Science
Foundation of Jiangsu Province (BK20220819),  and the Fundamental Research
Funds for the Central Universities (2242024k30035). This research work is also supported by the Big
Data Computing Center of Southeast University.
{
    \small
    \bibliographystyle{ieeenat_fullname}
    \bibliography{main}

\begin{thebibliography}{59}
\providecommand{\natexlab}[1]{#1}
\providecommand{\url}[1]{\texttt{#1}}
\expandafter\ifx\csname urlstyle\endcsname\relax
  \providecommand{\doi}[1]{doi: #1}\else
  \providecommand{\doi}{doi: \begingroup \urlstyle{rm}\Url}\fi

\bibitem[Alayrac et~al.(2022)Alayrac, Donahue, Luc, Miech, Barr, Hasson, Lenc, Mensch, Millican, Reynolds, et~al.]{flamingo}
Jean-Baptiste Alayrac, Jeff Donahue, Pauline Luc, Antoine Miech, Iain Barr, Yana Hasson, Karel Lenc, Arthur Mensch, Katherine Millican, Malcolm Reynolds, et~al.
\newblock Flamingo: a visual language model for few-shot learning.
\newblock \emph{Advances in neural information processing systems}, 35:\penalty0 23716--23736, 2022.

\bibitem[Baldassini et~al.(2024)Baldassini, Shukor, Cord, Soulier, and Piwowarski]{baldassini2024makes}
Folco~Bertini Baldassini, Mustafa Shukor, Matthieu Cord, Laure Soulier, and Benjamin Piwowarski.
\newblock What makes multimodal in-context learning work?
\newblock In \emph{Proceedings of the IEEE/CVF Conference on Computer Vision and Pattern Recognition}, pages 1539--1550, 2024.

\bibitem[Brown et~al.(2020)Brown, Mann, Ryder, Subbiah, Kaplan, Dhariwal, Neelakantan, Shyam, Sastry, Askell, Agarwal, Herbert-Voss, Krueger, Henighan, Child, Ramesh, Ziegler, Wu, Winter, Hesse, Chen, Sigler, Litwin, Gray, Chess, Clark, Berner, McCandlish, Radford, Sutskever, and Amodei]{gpt3}
Tom Brown, Benjamin Mann, Nick Ryder, Melanie Subbiah, Jared~D Kaplan, Prafulla Dhariwal, Arvind Neelakantan, Pranav Shyam, Girish Sastry, Amanda Askell, Sandhini Agarwal, Ariel Herbert-Voss, Gretchen Krueger, Tom Henighan, Rewon Child, Aditya Ramesh, Daniel Ziegler, Jeffrey Wu, Clemens Winter, Chris Hesse, Mark Chen, Eric Sigler, Mateusz Litwin, Scott Gray, Benjamin Chess, Jack Clark, Christopher Berner, Sam McCandlish, Alec Radford, Ilya Sutskever, and Dario Amodei.
\newblock Language models are few-shot learners.
\newblock In \emph{Advances in Neural Information Processing Systems}, pages 1877--1901. Curran Associates, Inc., 2020.

\bibitem[Cao et~al.(2022)Cao, Li, Li, Nie, and Zhang]{cao2022image}
Min Cao, Shiping Li, Juntao Li, Liqiang Nie, and Min Zhang.
\newblock Image-text retrieval: A survey on recent research and development.
\newblock \emph{arXiv preprint arXiv:2203.14713}, 2022.

\bibitem[Chen et~al.(2015)Chen, Fang, Lin, Vedantam, Gupta, Doll{\'a}r, and Zitnick]{cococap}
Xinlei Chen, Hao Fang, Tsung-Yi Lin, Ramakrishna Vedantam, Saurabh Gupta, Piotr Doll{\'a}r, and C~Lawrence Zitnick.
\newblock Microsoft coco captions: Data collection and evaluation server.
\newblock \emph{arXiv preprint arXiv:1504.00325}, 2015.

\bibitem[Dai et~al.(2022)Dai, Sun, Dong, Hao, Ma, Sui, and Wei]{icl-grad-descent}
Damai Dai, Yutao Sun, Li Dong, Yaru Hao, Shuming Ma, Zhifang Sui, and Furu Wei.
\newblock Why can gpt learn in-context? language models implicitly perform gradient descent as meta-optimizers.
\newblock \emph{arXiv preprint arXiv:2212.10559}, 2022.

\bibitem[Dong et~al.(2022)Dong, Li, Dai, Zheng, Ma, Li, Xia, Xu, Wu, Liu, et~al.]{iclsurvey}
Qingxiu Dong, Lei Li, Damai Dai, Ce Zheng, Jingyuan Ma, Rui Li, Heming Xia, Jingjing Xu, Zhiyong Wu, Tianyu Liu, et~al.
\newblock A survey on in-context learning.
\newblock \emph{arXiv preprint arXiv:2301.00234}, 2022.

\bibitem[Fu et~al.(2023)Fu, Chen, Shen, Qin, Zhang, Lin, Qiu, Lin, Yang, Zheng, Li, Sun, and Ji]{mme}
Chaoyou Fu, Peixian Chen, Yunhang Shen, Yulei Qin, Mengdan Zhang, Xu Lin, Zhenyu Qiu, Wei Lin, Jinrui Yang, Xiawu Zheng, Ke Li, Xing Sun, and Rongrong Ji.
\newblock Mme: A comprehensive evaluation benchmark for multimodal large language models.
\newblock \emph{ArXiv}, abs/2306.13394, 2023.

\bibitem[Gao et~al.(2021)Gao, Fisch, and Chen]{gao-etal-2021-making}
Tianyu Gao, Adam Fisch, and Danqi Chen.
\newblock Making pre-trained language models better few-shot learners.
\newblock In \emph{Proceedings of the 59th Annual Meeting of the Association for Computational Linguistics and the 11th International Joint Conference on Natural Language Processing (Volume 1: Long Papers)}, pages 3816--3830, Online, 2021. Association for Computational Linguistics.

\bibitem[Goyal et~al.(2017)Goyal, Khot, Summers-Stay, Batra, and Parikh]{vqav2}
Yash Goyal, Tejas Khot, Douglas Summers-Stay, Dhruv Batra, and Devi Parikh.
\newblock Making the v in vqa matter: Elevating the role of image understanding in visual question answering.
\newblock In \emph{Proceedings of the IEEE conference on computer vision and pattern recognition}, pages 6904--6913, 2017.

\bibitem[Guo et~al.(2024)Guo, Wang, Zhang, Chin, Liu, Cheng, Pan, Lee, Xue, Shen, et~al.]{guo2024scaling}
Wei Guo, Hao Wang, Luankang Zhang, Jin~Yao Chin, Zhongzhou Liu, Kai Cheng, Qiushi Pan, Yi~Quan Lee, Wanqi Xue, Tingjia Shen, et~al.
\newblock Scaling new frontiers: Insights into large recommendation models.
\newblock \emph{arXiv preprint arXiv:2412.00714}, 2024.

\bibitem[He et~al.(2023)He, Yang, Ge, Chen, Coatrieux, Wang, and Li]{he2023geometric}
Yuting He, Guanyu Yang, Rongjun Ge, Yang Chen, Jean-Louis Coatrieux, Boyu Wang, and Shuo Li.
\newblock Geometric visual similarity learning in 3d medical image self-supervised pre-training.
\newblock In \emph{Proceedings of the IEEE/CVF Conference on Computer Vision and Pattern Recognition}, pages 9538--9547, 2023.

\bibitem[Hendel et~al.(2023)Hendel, Geva, and Globerson]{taskvector}
Roee Hendel, Mor Geva, and Amir Globerson.
\newblock In-context learning creates task vectors.
\newblock \emph{arXiv preprint arXiv:2310.15916}, 2023.

\bibitem[Hu et~al.(2021)Hu, Shen, Wallis, Allen-Zhu, Li, Wang, Wang, and Chen]{lora}
Edward~J Hu, Yelong Shen, Phillip Wallis, Zeyuan Allen-Zhu, Yuanzhi Li, Shean Wang, Lu Wang, and Weizhu Chen.
\newblock Lora: Low-rank adaptation of large language models.
\newblock \emph{arXiv preprint arXiv:2106.09685}, 2021.

\bibitem[Huang et~al.(2024)Huang, Mitra, Arbelle, Karlinsky, Darrell, and Herzig]{mtv}
Brandon Huang, Chancharik Mitra, Assaf Arbelle, Leonid Karlinsky, Trevor Darrell, and Roei Herzig.
\newblock Multimodal task vectors enable many-shot multimodal in-context learning.
\newblock \emph{arXiv preprint arXiv:2406.15334}, 2024.

\bibitem[Jiang et~al.(2024)Jiang, Zhou, Li, Lu, Wang, Ma, Chang, and Zhang]{jiang2024dgpic}
Jincen Jiang, Qianyu Zhou, Yuhang Li, Xuequan Lu, Meili Wang, Lizhuang Ma, Jian Chang, and Jian~Jun Zhang.
\newblock Dg-pic: Domain generalized point-in-context learning for point cloud understanding.
\newblock In \emph{European Conference on Computer Vision (ECCV)}, pages 455--474. Springer, 2024.

\bibitem[Kumar and Talukdar(2021)]{kumar2021reordering}
Sawan Kumar and Partha Talukdar.
\newblock Reordering examples helps during priming-based few-shot learning.
\newblock \emph{arXiv preprint arXiv:2106.01751}, 2021.

\bibitem[Lauren{\c{c}}on et~al.(2024{\natexlab{a}})Lauren{\c{c}}on, Saulnier, Tronchon, Bekman, Singh, Lozhkov, Wang, Karamcheti, Rush, Kiela, et~al.]{idefics1}
Hugo Lauren{\c{c}}on, Lucile Saulnier, L{\'e}o Tronchon, Stas Bekman, Amanpreet Singh, Anton Lozhkov, Thomas Wang, Siddharth Karamcheti, Alexander Rush, Douwe Kiela, et~al.
\newblock Obelics: An open web-scale filtered dataset of interleaved image-text documents.
\newblock \emph{Advances in Neural Information Processing Systems}, 36, 2024{\natexlab{a}}.

\bibitem[Lauren{\c{c}}on et~al.(2024{\natexlab{b}})Lauren{\c{c}}on, Tronchon, Cord, and Sanh]{idefics2}
Hugo Lauren{\c{c}}on, L{\'e}o Tronchon, Matthieu Cord, and Victor Sanh.
\newblock What matters when building vision-language models?
\newblock \emph{arXiv preprint arXiv:2405.02246}, 2024{\natexlab{b}}.

\bibitem[Li et~al.(2024{\natexlab{a}})Li, Ge, Ge, Wang, Wang, Zhang, and Shan]{seed}
Bohao Li, Yuying Ge, Yixiao Ge, Guangzhi Wang, Rui Wang, Ruimao Zhang, and Ying Shan.
\newblock Seed-bench: Benchmarking multimodal large language models.
\newblock In \emph{Proceedings of the IEEE/CVF Conference on Computer Vision and Pattern Recognition (CVPR)}, pages 13299--13308, 2024{\natexlab{a}}.

\bibitem[Li et~al.(2024{\natexlab{b}})Li, Zhang, Guo, Zhang, Li, Zhang, Zhang, Li, Liu, and Li]{li2024llava-one}
Bo Li, Yuanhan Zhang, Dong Guo, Renrui Zhang, Feng Li, Hao Zhang, Kaichen Zhang, Yanwei Li, Ziwei Liu, and Chunyuan Li.
\newblock Llava-onevision: Easy visual task transfer.
\newblock \emph{arXiv preprint arXiv:2408.03326}, 2024{\natexlab{b}}.

\bibitem[Li et~al.(2024{\natexlab{c}})Li, Zhang, Zhang, Zhang, Li, Li, Ma, and Li]{li2024llava}
Feng Li, Renrui Zhang, Hao Zhang, Yuanhan Zhang, Bo Li, Wei Li, Zejun Ma, and Chunyuan Li.
\newblock Llava-next-interleave: Tackling multi-image, video, and 3d in large multimodal models.
\newblock \emph{arXiv preprint arXiv:2407.07895}, 2024{\natexlab{c}}.

\bibitem[Li et~al.(2024{\natexlab{d}})Li, Peng, Chen, Gao, and Yang]{cfg-icd}
Li Li, Jiawei Peng, Huiyi Chen, Chongyang Gao, and Xu Yang.
\newblock How to configure good in-context sequence for visual question answering.
\newblock In \emph{Proceedings of the IEEE/CVF Conference on Computer Vision and Pattern Recognition}, pages 26710--26720, 2024{\natexlab{d}}.

\bibitem[Li et~al.(2023)Li, Lv, Yan, Lin, Zhu, Ni, Xie, Wang, and Qiu]{li2023unified}
Xiaonan Li, Kai Lv, Hang Yan, Tianyang Lin, Wei Zhu, Yuan Ni, Guotong Xie, Xiaoling Wang, and Xipeng Qiu.
\newblock Unified demonstration retriever for in-context learning.
\newblock \emph{arXiv preprint arXiv:2305.04320}, 2023.

\bibitem[Liu et~al.(2024)Liu, Li, Wu, and Lee]{llava}
Haotian Liu, Chunyuan Li, Qingyang Wu, and Yong~Jae Lee.
\newblock Visual instruction tuning.
\newblock \emph{Advances in neural information processing systems}, 36, 2024.

\bibitem[Liu et~al.(2022)Liu, Shen, Zhang, Dolan, Carin, and Chen]{icd-sim-selection}
Jiachang Liu, Dinghan Shen, Yizhe Zhang, Bill Dolan, Lawrence Carin, and Weizhu Chen.
\newblock What makes good in-context examples for {GPT}-3?
\newblock In \emph{Proceedings of Deep Learning Inside Out (DeeLIO 2022): The 3rd Workshop on Knowledge Extraction and Integration for Deep Learning Architectures}, pages 100--114, Dublin, Ireland and Online, 2022. Association for Computational Linguistics.

\bibitem[Liu et~al.(2023)Liu, Ye, Xing, and Zou]{icv}
Sheng Liu, Haotian Ye, Lei Xing, and James Zou.
\newblock In-context vectors: Making in context learning more effective and controllable through latent space steering.
\newblock \emph{arXiv preprint arXiv:2311.06668}, 2023.

\bibitem[Lu et~al.(2022)Lu, Bartolo, Moore, Riedel, and Stenetorp]{order-sensitivity}
Yao Lu, Max Bartolo, Alastair Moore, Sebastian Riedel, and Pontus Stenetorp.
\newblock Fantastically ordered prompts and where to find them: Overcoming few-shot prompt order sensitivity.
\newblock In \emph{Proceedings of the 60th Annual Meeting of the Association for Computational Linguistics (Volume 1: Long Papers)}, pages 8086--8098, Dublin, Ireland, 2022. Association for Computational Linguistics.

\bibitem[Luo et~al.(2024{\natexlab{a}})Luo, Xu, Liu, Pasupat, and Kazemi]{luo2024context}
Man Luo, Xin Xu, Yue Liu, Panupong Pasupat, and Mehran Kazemi.
\newblock In-context learning with retrieved demonstrations for language models: A survey.
\newblock \emph{arXiv preprint arXiv:2401.11624}, 2024{\natexlab{a}}.

\bibitem[Luo et~al.(2024{\natexlab{b}})Luo, Zheng, Zhu, and You]{luo2024does}
Yang Luo, Zangwei Zheng, Zirui Zhu, and Yang You.
\newblock How does the textual information affect the retrieval of multimodal in-context learning?
\newblock \emph{arXiv preprint arXiv:2404.12866}, 2024{\natexlab{b}}.

\bibitem[Marino et~al.(2019)Marino, Rastegari, Farhadi, and Mottaghi]{okvqa}
Kenneth Marino, Mohammad Rastegari, Ali Farhadi, and Roozbeh Mottaghi.
\newblock Ok-vqa: A visual question answering benchmark requiring external knowledge.
\newblock In \emph{Proceedings of the IEEE/cvf conference on computer vision and pattern recognition}, pages 3195--3204, 2019.

\bibitem[Min et~al.(2022)Min, Lyu, Holtzman, Artetxe, Lewis, Hajishirzi, and Zettlemoyer]{rethink-icd-role}
Sewon Min, Xinxi Lyu, Ari Holtzman, Mikel Artetxe, Mike Lewis, Hannaneh Hajishirzi, and Luke Zettlemoyer.
\newblock Rethinking the role of demonstrations: What makes in-context learning work?
\newblock \emph{arXiv preprint arXiv:2202.12837}, 2022.

\bibitem[Mosbach et~al.(2023)Mosbach, Pimentel, Ravfogel, Klakow, and Elazar]{ft-vs-icl}
Marius Mosbach, Tiago Pimentel, Shauli Ravfogel, Dietrich Klakow, and Yanai Elazar.
\newblock Few-shot fine-tuning vs. in-context learning: A fair comparison and evaluation.
\newblock \emph{arXiv preprint arXiv:2305.16938}, 2023.

\bibitem[Peng et~al.(2023)Peng, Yang, Ma, Xu, Zhang, Han, and Zhang]{lever-lm}
Yingzhe Peng, Xu Yang, Haoxuan Ma, Shuo Xu, Chi Zhang, Yucheng Han, and Hanwang Zhang.
\newblock Icd-lm: Configuring vision-language in-context demonstrations by language modeling.
\newblock \emph{arXiv preprint arXiv:2312.10104}, 2023.

\bibitem[Peng et~al.(2024)Peng, Hao, Yang, Peng, Hu, and Geng]{licv}
Yingzhe Peng, Chenduo Hao, Xu Yang, Jiawei Peng, Xinting Hu, and Xin Geng.
\newblock Learnable in-context vector for visual question answering.
\newblock \emph{arXiv preprint arXiv:2406.13185}, 2024.

\bibitem[Peng et~al.(2025)Peng, Zhang, Zhang, You, Liu, Zhu, Yang, Xu, Geng, and Yang]{peng2025lmm}
Yingzhe Peng, Gongrui Zhang, Miaosen Zhang, Zhiyuan You, Jie Liu, Qipeng Zhu, Kai Yang, Xingzhong Xu, Xin Geng, and Xu Yang.
\newblock Lmm-r1: Empowering 3b lmms with strong reasoning abilities through two-stage rule-based rl.
\newblock \emph{arXiv preprint arXiv:2503.07536}, 2025.

\bibitem[Radford et~al.(2019)Radford, Wu, Child, Luan, Amodei, Sutskever, et~al.]{radford2019language}
Alec Radford, Jeffrey Wu, Rewon Child, David Luan, Dario Amodei, Ilya Sutskever, et~al.
\newblock Language models are unsupervised multitask learners.
\newblock 2019.

\bibitem[Rohrbach et~al.(2018)Rohrbach, Hendricks, Burns, Darrell, and Saenko]{chair}
Anna Rohrbach, Lisa~Anne Hendricks, Kaylee Burns, Trevor Darrell, and Kate Saenko.
\newblock Object hallucination in image captioning.
\newblock \emph{arXiv preprint arXiv:1809.02156}, 2018.

\bibitem[Rubin et~al.(2022)Rubin, Herzig, and Berant]{rubin-etal-2022-learning}
Ohad Rubin, Jonathan Herzig, and Jonathan Berant.
\newblock Learning to retrieve prompts for in-context learning.
\newblock In \emph{Proceedings of the 2022 Conference of the North American Chapter of the Association for Computational Linguistics: Human Language Technologies}, pages 2655--2671, Seattle, United States, 2022. Association for Computational Linguistics.

\bibitem[Shen et~al.(2024)Shen, Wang, Wu, Chin, Guo, Liu, Guo, Lian, Tang, and Chen]{shen2024optimizing}
Tingjia Shen, Hao Wang, Chuhan Wu, Jin~Yao Chin, Wei Guo, Yong Liu, Huifeng Guo, Defu Lian, Ruiming Tang, and Enhong Chen.
\newblock Optimizing sequential recommendation models with scaling laws and approximate entropy.
\newblock \emph{arXiv preprint arXiv:2412.00430}, 2024.

\bibitem[Shukor et~al.(2023)Shukor, Rame, Dancette, and Cord]{shukor2023beyond}
Mustafa Shukor, Alexandre Rame, Corentin Dancette, and Matthieu Cord.
\newblock Beyond task performance: Evaluating and reducing the flaws of large multimodal models with in-context learning.
\newblock \emph{arXiv preprint arXiv:2310.00647}, 2023.

\bibitem[Sun et~al.(2024)Sun, Zhou, Li, Lu, Yi, Chen, Xu, Luo, Zhang, Zhan, et~al.]{parrot}
Hai-Long Sun, Da-Wei Zhou, Yang Li, Shiyin Lu, Chao Yi, Qing-Guo Chen, Zhao Xu, Weihua Luo, Kaifu Zhang, De-Chuan Zhan, et~al.
\newblock Parrot: Multilingual visual instruction tuning.
\newblock \emph{arXiv preprint arXiv:2406.02539}, 2024.

\bibitem[Tanwar et~al.(2023)Tanwar, Dutta, Borthakur, and Chakraborty]{icd-sim-selection2}
Eshaan Tanwar, Subhabrata Dutta, Manish Borthakur, and Tanmoy Chakraborty.
\newblock Multilingual {LLM}s are better cross-lingual in-context learners with alignment.
\newblock In \emph{Proceedings of the 61st Annual Meeting of the Association for Computational Linguistics (Volume 1: Long Papers)}, pages 6292--6307, Toronto, Canada, 2023. Association for Computational Linguistics.

\bibitem[Todd et~al.(2023)Todd, Li, Sharma, Mueller, Wallace, and Bau]{functionvector}
Eric Todd, Millicent~L Li, Arnab~Sen Sharma, Aaron Mueller, Byron~C Wallace, and David Bau.
\newblock Function vectors in large language models.
\newblock \emph{arXiv preprint arXiv:2310.15213}, 2023.

\bibitem[Wang et~al.(2024)Wang, Bai, Tan, Wang, Fan, Bai, Chen, Liu, Wang, Ge, et~al.]{wang2024qwen2}
Peng Wang, Shuai Bai, Sinan Tan, Shijie Wang, Zhihao Fan, Jinze Bai, Keqin Chen, Xuejing Liu, Jialin Wang, Wenbin Ge, et~al.
\newblock Qwen2-vl: Enhancing vision-language model's perception of the world at any resolution.
\newblock \emph{arXiv preprint arXiv:2409.12191}, 2024.

\bibitem[Wei et~al.(2022)Wei, Wang, Schuurmans, Bosma, ichter, Xia, Chi, Le, and Zhou]{cot-reasoning}
Jason Wei, Xuezhi Wang, Dale Schuurmans, Maarten Bosma, brian ichter, Fei Xia, Ed Chi, Quoc~V Le, and Denny Zhou.
\newblock Chain-of-thought prompting elicits reasoning in large language models.
\newblock In \emph{Advances in Neural Information Processing Systems}, pages 24824--24837. Curran Associates, Inc., 2022.

\bibitem[Wu et~al.(2023)Wu, Wang, Ye, and Kong]{wu-etal-2023-self}
Zhiyong Wu, Yaoxiang Wang, Jiacheng Ye, and Lingpeng Kong.
\newblock Self-adaptive in-context learning: An information compression perspective for in-context example selection and ordering.
\newblock In \emph{Proceedings of the 61st Annual Meeting of the Association for Computational Linguistics (Volume 1: Long Papers)}, pages 1423--1436, Toronto, Canada, 2023. Association for Computational Linguistics.

\bibitem[Xiao et~al.(2023)Xiao, Tian, Chen, Han, and Lewis]{long-ctx-cost-time}
Guangxuan Xiao, Yuandong Tian, Beidi Chen, Song Han, and Mike Lewis.
\newblock Efficient streaming language models with attention sinks.
\newblock \emph{arXiv preprint arXiv:2309.17453}, 2023.

\bibitem[Xu et~al.(2024)Xu, Wang, Zhang, Poon, and Chen]{xu2024introspection}
Nan Xu, Fei Wang, Sheng Zhang, Hoifung Poon, and Muhao Chen.
\newblock From introspection to best practices: Principled analysis of demonstrations in multimodal in-context learning.
\newblock \emph{arXiv preprint arXiv:2407.00902}, 2024.

\bibitem[Yang et~al.(2024)Yang, Wu, Yang, Chen, and Geng]{wyl}
Xu Yang, Yongliang Wu, Mingzhuo Yang, Haokun Chen, and Xin Geng.
\newblock Exploring diverse in-context configurations for image captioning.
\newblock \emph{Advances in Neural Information Processing Systems}, 36, 2024.

\bibitem[Yang et~al.(2022)Yang, Gan, Wang, Hu, Lu, Liu, and Wang]{rices}
Zhengyuan Yang, Zhe Gan, Jianfeng Wang, Xiaowei Hu, Yumao Lu, Zicheng Liu, and Lijuan Wang.
\newblock An empirical study of gpt-3 for few-shot knowledge-based vqa.
\newblock In \emph{Proceedings of the AAAI conference on artificial intelligence}, pages 3081--3089, 2022.

\bibitem[Ye et~al.(2023)Ye, Wu, Feng, Yu, and Kong]{ye2023compositional}
Jiacheng Ye, Zhiyong Wu, Jiangtao Feng, Tao Yu, and Lingpeng Kong.
\newblock Compositional exemplars for in-context learning.
\newblock In \emph{International Conference on Machine Learning}, pages 39818--39833. PMLR, 2023.

\bibitem[Yi et~al.(2024)Yi, He, Zhan, and Ye]{yi2024bridge}
Chao Yi, Yuhang He, De-Chuan Zhan, and Han-Jia Ye.
\newblock Bridge the modality and capability gaps in vision-language model selection.
\newblock \emph{Advances in Neural Information Processing Systems}, 37:\penalty0 34429--34452, 2024.

\bibitem[Yin et~al.(2024)Yin, Wang, Guo, Liu, Zhang, Zhao, Lian, and Chen]{yin2024dataset}
Mingjia Yin, Hao Wang, Wei Guo, Yong Liu, Suojuan Zhang, Sirui Zhao, Defu Lian, and Enhong Chen.
\newblock Dataset regeneration for sequential recommendation.
\newblock In \emph{Proceedings of the 30th ACM SIGKDD Conference on Knowledge Discovery and Data Mining}, pages 3954--3965, 2024.

\bibitem[Yin et~al.(2023)Yin, Fu, Zhao, Li, Sun, Xu, and Chen]{yin2023survey}
Shukang Yin, Chaoyou Fu, Sirui Zhao, Ke Li, Xing Sun, Tong Xu, and Enhong Chen.
\newblock A survey on multimodal large language models.
\newblock \emph{arXiv preprint arXiv:2306.13549}, 2023.

\bibitem[Young et~al.(2014)Young, Lai, Hodosh, and Hockenmaier]{flickr30k}
Peter Young, Alice Lai, Micah Hodosh, and Julia Hockenmaier.
\newblock From image descriptions to visual denotations: New similarity metrics for semantic inference over event descriptions.
\newblock \emph{Transactions of the Association for Computational Linguistics}, 2:\penalty0 67--78, 2014.

\bibitem[Yue et~al.(2024)Yue, Zheng, Ni, Wang, Zhang, Tong, Sun, Yin, Yu, Zhang, Sun, Su, Chen, and Neubig]{mmmu-pro}
Xiang Yue, Tianyu Zheng, Yuansheng Ni, Yubo Wang, Kai Zhang, Shengbang Tong, Yuxuan Sun, Ming Yin, Botao Yu, Ge Zhang, Huan Sun, Yu Su, Wenhu Chen, and Graham Neubig.
\newblock Mmmu-pro: A more robust multi-discipline multimodal understanding benchmark.
\newblock \emph{ArXiv}, abs/2409.02813, 2024.

\bibitem[Zhao et~al.(2021)Zhao, Wallace, Feng, Klein, and Singh]{cali-before-use}
Zihao Zhao, Eric Wallace, Shi Feng, Dan Klein, and Sameer Singh.
\newblock Calibrate before use: Improving few-shot performance of language models.
\newblock In \emph{Proceedings of the 38th International Conference on Machine Learning}, pages 12697--12706. PMLR, 2021.

\bibitem[Zhao et~al.(2024)Zhao, Tang, Lin, Wu, Huang, Liu, Tan, Zhang, and Xie]{zhao2024multi}
Zhen Zhao, Jingqun Tang, Chunhui Lin, Binghong Wu, Can Huang, Hao Liu, Xin Tan, Zhizhong Zhang, and Yuan Xie.
\newblock Multi-modal in-context learning makes an ego-evolving scene text recognizer.
\newblock In \emph{Proceedings of the IEEE/CVF Conference on Computer Vision and Pattern Recognition}, pages 15567--15576, 2024.

\end{thebibliography}
}

\clearpage
\setcounter{page}{1}
\maketitlesupplementary
\section{Implementation Details}
\subsection{Prompts}
Following \cite{idefics1}, we use the same prompts for both Idefics1 and Idefics2, as shown in ~\cref{tab:prompts}.
\begin{table*}[th]
\centering
\resizebox{\textwidth}{!}{
\begin{tabular}{@{}cccc@{}}
\toprule
\textbf{Task} & \textbf{Prefix prompt} & \textbf{ICD prompt} & \textbf{Stop words} \\ \midrule
\makecell{VQAv2 \\ OK-VQA} & \makecell[l]{Instruction: provide an answer \\ to the question. Use the image to answer.} & \makecell[l]{Image:\{image\} Question: \{question\} \\ Answer: \{answer\}\textbackslash{}n} & ``Question'', ``Answer'',``Image'' \\ \midrule
\makecell{COCO} & \XSolidBrush & \multirow{2}{*}{Image: \{image\} Caption: \{caption\}\textbackslash{}n} & \multirow{2}{*}{``Caption'', ``Image''} \\ \cmidrule(r){1-2}
\multicolumn{1}{l}{\makecell{COCO ICL}} & \multicolumn{1}{l}{\makecell[l]{Instruction: provide a short caption \\ of the input image.\textbackslash{}n}} &  &  \\ \bottomrule
\end{tabular}
}
\caption{The prompt templates on different tasks evaluated in our paper. The data to be replaced is between curly brackets.}
\label{tab:prompts}
\end{table*}
\subsection{Hyperparameters}
For all datasets, the hyperparameters for all trainable methods are as outlined in \cref{tab:hyper-params}. When the training set size is less than 1000, we perform training for 10 epochs, and when the training set size is 1000, we perform training for 5 epochs. For LIVE, we follow the recommendations from the original paper and train for 10 epochs when the training set size is 1000~\cite{licv}. For both LIVE and MimIC, we use the same learning rate of $5 \times 10^{-3}$, and the learning rate for the shift magnitude in LIVE is set to $1 \times 10^{-2}$, in accordance with the original paper.

For LoRA, we set the rank $r=16$ and modified $\bm{W}_q$, $\bm{W}_k$, $\bm{W}_v$, and $\bm{W}_o$ in all attention layers of both the vision and language models. Given the substantial number of parameters in LoRA, we set the learning rate to $5 \times 10^{-4}$ to ensure stable training. In \cref{sec:more-analyze}, we introduce an alternative parameter setting for LoRA, denoted as LoRA$^\dagger$. Specifically, LoRA$^\dagger$ modifies only $\bm{W}_o$ in the language model, with a rank of $r=1$. This configuration is the most similar to MimIC, not only in terms of the number of parameters but also in the modification of the input to the feed-forward network (FFN) layer. For both configurations, the dropout rate is set to 0.05, and the LoRA scaling factor $\alpha$ is set to $2r$.
\begin{table}[h]
\centering
\begin{tabular}{@{}cc@{}}
\toprule
\multicolumn{1}{c}{\textbf{Hyperparameters}} & \textbf{Value} \\ \midrule
optimizer & AdamW \\
warmup step ratio & 0.1 \\
precision & float16 \\
weight decay & 1e-3 \\
batch size & 2 \\
accumulate gradient batches & 2 \\ \bottomrule
\end{tabular}
\caption{The common hyperparameters used in training for all trainable methods over all datasets on both Idefics1 and Idefics2. }
\label{tab:hyper-params}
\end{table}
\section{Exploratory Experiments}
\subsection{Where to Align?}
\begin{figure*}[th]
    \centering
    \includegraphics[width=0.3\linewidth]{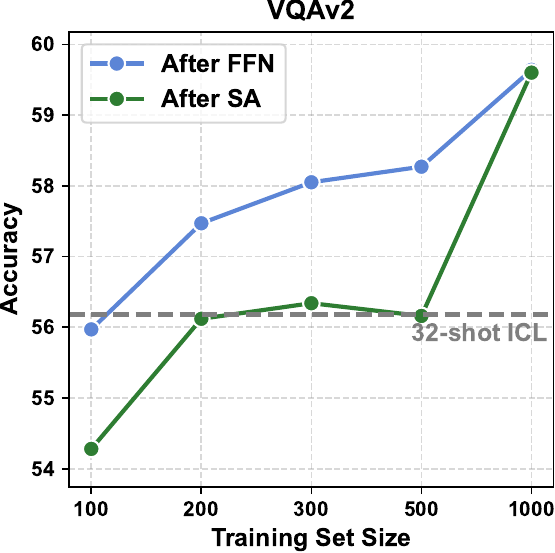}
    \includegraphics[width=0.3\linewidth]{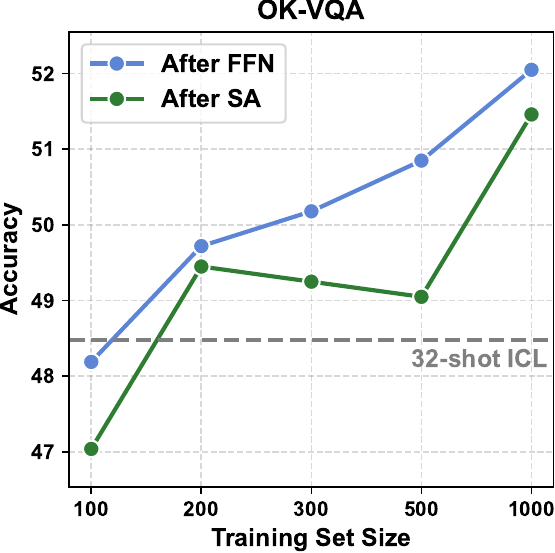}
    \includegraphics[width=0.3\linewidth]{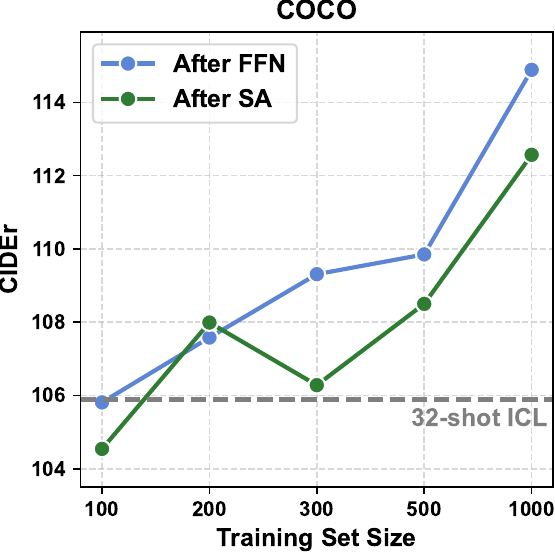}
    \caption{Performance of MimIC trained with different alignment strategy.}
    \label{fig:align-pos}
\end{figure*}
   
In~\cref{sec:mimic}, we show that MimIC uses the output of the FFN layer in each decoder layer of both the original LMM and the MimIC LMM to compute $\mathcal{L}_{\textrm{align}}$, aiming to align zero-shot and ICL. However, using the self-attention output to compute $\mathcal{L}_{\textrm{align}}$ is also reasonable, as~\cref{eq:expand-head} only requires that the shift vector should be added after the self-attention. Therefore, on Idefics1, we evaluate two settings: (1) \textbf{After SA}: using the hidden states from the self-attention output; (2) \textbf{After FFN}: using the hidden states from the FFN output, which is adopted by MimIC. The results are presented in~\cref{fig:align-pos}. We find that while ``After SA'' converges faster, its performance is inferior to ``After FFN'', especially when the training set size is small. This may be because the FFN amplifies the errors in the attention output, making it easier for the MimIC attention head to overfit and leading to poorer generalization performance.
\subsection{Implementations of the Shift Vector}

In \cref{sec:formula}, we decompose the single-head self-attention (SA) for each query in ICL into the following components: standard attention $SA(\bm{q}, \bm{K}, \bm{V})$, shift magnitude $\mu$, and the attention difference term $\textrm{SA}(\bm{q}, \bm{K}_D, \bm{V}_D) - \textrm{SA}(\bm{q}, \bm{K}, \bm{V})$. Moreover, \cref{eq:expand-head} can be rewritten as:

\begin{equation}
    \begin{aligned}
        &\textrm{SA}\left(\bm{q}, \begin{bmatrix} \bm{K}_D \\ \bm{K} \end{bmatrix}, \begin{bmatrix} \bm{V}_D \\ \bm{V} \end{bmatrix}\right) \\
    &= \textrm{SA}(\bm{q}, \bm{K}, \bm{V}) + \mu \left( \textrm{SA}(\bm{q}, \bm{K}_D, \bm{V}_D) - \textrm{SA}(\bm{q}, \bm{K}, \bm{V}) \right) \\
    &= (1-\mu) \underbrace{\textrm{SA}(\bm{q}, \bm{K}, \bm{V})}_{\textrm{standard attention}} + \mu \underbrace{\textrm{SA}(\bm{q}, \bm{K}_D, \bm{V}_D)}_{\textrm{attention over ICDs}}
    \end{aligned}
\end{equation}

Note that the attention over ICDs in the second term depends solely on ICDs and is independent of other query tokens. Therefore, we can approximate $\textrm{SA}(\bm{q}, \bm{K}_D, \bm{V}_D)$ using a network, \ie, $h(q) := \textrm{SA}(\bm{q}, \bm{K}_D, \bm{V}_D)$. For simplicity, we train a linear layer $h: \mathbb{R}^{d_h} \rightarrow \mathbb{R}^{d_h}$. The results, shown in \cref{tab:shift-impl}, indicate that using a linear layer to implement the shift vector performs significantly worse than MimIC. This may be due to the query-dependent shift being more sensitive to noise from different ICD configurations, making it less robust than using a query-independent learnable vector. To verify this, we replace the linear layer $h$ with the learnable vector, and observe a substantial improvement in performance.

\begin{table}[th]
\centering
\begin{tabular}{@{}cccc@{}}
\toprule
 \textbf{Method} & \textbf{VQAv2} & OK-\textbf{VQA} & \textbf{COCO} \\ \midrule
Linear layer & 47.68 & 42.61 & 112.88 \\
Learnable vector & 58.84 & 51.13 & 113.07 \\
MimIC & \textbf{59.64} & \textbf{52.05} & \textbf{114.89} \\ \bottomrule
\end{tabular}
\caption{Performance comparison among different implementations of the shift vector.}
\label{tab:shift-impl}
\end{table}
\section{Additional Results}
\subsection{Training with Scaling Data}
In \cref{sec:main-result}, we evaluated the performance of MimIC on 1000 data samples and also examined the effect of reducing the number of training samples. However, we did not investigate how MimIC performs when scaling up the dataset. A limited number of samples may not be sufficient for MimIC to reach optimal performance. Therefore, we further validated its performance on 8000 samples.

As shown in \cref{tab:scaling-data}, MimIC consistently outperforms LoRA in most cases and remains ahead of LIVE. While its performance improves with a larger dataset, the performance gap between MimIC and LoRA narrows and, in some cases, is even reversed. This phenomenon occurs because MimIC is trained by simulating in-context learning (ICL), which imposes inherent constraints on its upper performance limit. MimIC was designed for the low-data regime; however, if this limitation can be mitigated, MimIC has the potential to become a highly efficient method for parameter-efficient fine-tuning.
\begin{table}[t]
\centering
\renewcommand\arraystretch{1.2}
\resizebox{\linewidth}{!}{
\begin{tabular}{@{}ccccc@{}}
\toprule
\textbf{Model} & \textbf{Method} & \textbf{VQAv2} & \textbf{OK-VQA} & \textbf{COCO} \\ \midrule
\multirow{5}{*}{\rotatebox{90}{Idefics-9b}} & Zero-shot & 29.25 & 30.54 & 63.06 \\
 & 32-shot ICL & 56.18 & 48.48 & 105.89 \\ \cmidrule[0.1pt]{2-5} 
 & LIVE & 58.54*& 50.08*& \underline{117.38}*\\
 & LoRA & \underline{59.04}& \underline{53.15}& 110.37\\
 & MimIC & \textbf{60.2}& \textbf{53.84}& \textbf{118.07}\\ \midrule
\multirow{5}{*}{\rotatebox{90}{Idefics2-8b-base}} & Zero-shot & 55.39 & 43.08 & 40.00 \\
 & 8-shot ICL & 66.20 & 57.68 & 122.51 \\ \cmidrule[0.1pt]{2-5} 
 & LIVE & 70.30*& 58.52*& -\\
 & LoRA & \textbf{75.24}& \textbf{64.26}& 133.8\\
 & MimIC & \underline{72.85}& \underline{61.76}& \textbf{133.98}\\ \bottomrule
\end{tabular}
}
\caption{
The results of VQAv2, OK-VQA, and COCO on Idefics-9b and Idefics2-8b-base trained on 8000 samples. The weight of alignment loss is set to 0.7. Numbers marked with an asterisk (*), in \textbf{bold}, or \underline{underlined} represent results reported in the original paper, the best results, and the second-best results, respectively. 
}

\label{tab:scaling-data}
\end{table}
\subsection{Generalize to more Tasks}
We evaluated MimIC on four new tasks: 1) \textbf{Flickr30k}~\cite{flickr30k}: A large image-caption dataset consisting of 31,000 images, each paired with five descriptive captions. 2) \textbf{MME}~\cite{mme}: A comprehensive benchmark designed to evaluate the performance of LMMs across 14 subtasks, assessing both perceptual and cognitive abilities. 3) \textbf{SEED-bench}~\cite{seed}: A novel benchmark comprising 24,000 multiple-choice questions with precise human annotations, covering 27 evaluation dimensions. 4) \textbf{MMMU-Pro}~\cite{mmmu-pro}: An advanced benchmark for evaluating large language models across multiple disciplines, featuring over 12,000 complex multiple-choice questions with ten options each, spanning 14 subjects such as mathematics, physics, and law.
The results are presented in ~\cref{tab:more-tasks}, which indicate that despite the increased difficulty of these benchmarks, our method consistently outperforms both LoRA and many-shot ICL in most cases.

Notably, due to computational resource constraints, we could only apply 2-shot ICL on MME, SEED-bench, and MMMU-Pro and were unable to train LoRA. In contrast, thanks to MimIC’s lightweight nature, it can still be successfully trained. Additionally, since training does not require storing the KV cache, we can even train MimIC using more-shot ICL. For instance, we can train MimIC with 16-shot on VQAv2, OK-VQA, and COCO, whereas ICL inference is limited to a maximum of 8-shot. This unique advantage also enables applications in scenarios where 1-shot ICL is not feasible.
\begin{table}[ht]
    \centering
    \renewcommand\arraystretch{1.4}
    \resizebox{\linewidth}{!}{
    \begin{tabular}{@{}cccccc@{}}
    \toprule
    \textbf{Model} & \textbf{Method} & \textbf{Flickr30k} & \textbf{MME} & \textbf{SEED}  &\textbf{MMMU-Pro} \\ \midrule
    \multirow{4}{*}{\rotatebox{90}{Idefics-9b}}& Zero-shot & 49.17 & 55.36 & 27.56&26.10\\
     & ICL& 63.41& 52.11& 28.30&28.14$^{16}$\\
     & LoRA & 72.79 & 60.53 & 26.95 &27.74\\
     & MimIC & \textbf{74.03} & \textbf{63.06} & \textbf{29.89}&\textbf{31.38}$^{16}$\\ \midrule
    \multirow{4}{*}{\rotatebox{90}{Idefics2-8b-base}}& Zero-shot & 53.04& 74.80& 12.91&28.92\\
     & ICL& 84.57& 71.10$^{2}$& \textbf{47.9}$^{2}$&\textbf{32.60}$^{2}$\\
     & LoRA & 73.03& -& -&-\\
     & MimIC & \textbf{91.77}& \textbf{80.83}$^{2}$& 47.00$^{2}$&31.73$^{2}$\\ \bottomrule
    \end{tabular}
    }
    \caption{Results evaluated on more tasks. The actual number of shots used is indicated in superscript. For cases without annotations, it is consistent with the description of \cref{sec:main-result}.}
    \label{tab:more-tasks}
\end{table}

\subsection{Effectiveness of MimIC in Alignment Effect}
\label{sec:more-analyze}
\begin{table}[t]
\centering
\begin{tabular}{@{}cccc@{}}
\toprule
 & \textbf{Method} & \textbf{VQAv2} & \textbf{OK-VQA} \\ \midrule
 & Zero-shot & 42.97 & 41.21 \\ \midrule
\multirow{2}{*}{without $\mathcal{L}_{\textrm{align}}$} & LoRA$^\dagger$ & 54.67 & 48.19 \\
 & LoRA & 61.18 & 47.18 \\ \midrule
\multirow{3}{*}{with $\mathcal{L}_{\textrm{align}}$} & LoRA$^\dagger$ & 37.32 & 36.77 \\
 & LoRA & 31.89 & 30.18 \\
 & MimIC & \textbf{30.17} & \textbf{28.24} \\ \bottomrule
\end{tabular}

\caption{Comparison of L2 distances between different LoRA settings and 32-shot ICL, and between MimIC and 32-shot ICL.  LoRA$^\dagger$ is a modified LoRA setting, which only modifies the output matrix $W_o$ of self attention layers in the language model with a rank $r=1$.}
\label{tab:align-lora}
\end{table}
Our quantitative analysis in~\cref{sec:ablation} demonstrates that the MimIC attention head and $\mathcal{L}_{\textrm{align}}$ outperform the shift vector and KL divergence used in LIVE, but no comparison has been made with LoRA. It's natural to ask: whether LoRA can mimic in-context learning in the MimIC framework, as both LoRA and MimIC add a small number of trainable parameters to LMMs.

To address this, similar to~\cref{sec:ablation}, we compute the average L2 distance of the latent representations of the first answer token at each layer, with or without $\mathcal{L}_{\textrm{align}}$ on LoRA, compared to the 32-shot ICL. For a more intuitive comparison, we also evaluate a modified LoRA setting, denoted as LoRA$^\dagger$, which only modifies the output matrix $W_o$ of self attention layers in the language model with a rank $r=1$. This setting ensures that the number of parameters in LoRA$^\dagger$ matches that of MimIC.

The results, presented in~\cref{tab:align-lora}, show that, regardless of whether LoRA has more parameters or the same number of parameters as MimIC, the distance from 32-shot ICL remains greater after training with $\mathcal{L}_{\textrm{align}}$. This indicates that the efficient design of the MimIC attention head allows it to more effectively mimic ICL.

\subsection{Marrying MimIC with LoRA}
Although MimIC is highly efficient, its limited number of parameters may restrict its capacity to learn more complex patterns. Fortunately, it is compatible with certain parameter efficient fine-tuning (PEFT) methods, making it possible to combine MimIC with various PEFT techniques. Here, we investigate the performance of integrating MimIC with LoRA. The results, shown in \cref{tab:mimic-lora}, indicate that adding LoRA leads to performance improvements for MimIC across three datasets, requiring only one epoch of training. This not only highlights the exceptional learning efficiency of MimIC but also suggests that using ICL as a guiding mechanism can enhance the adaptability of fine-tuning methods in few data scenarios. Furthermore, this combination may help bridge the performance gap between ICL and fine-tuning, allowing the two approaches to overcome the potential limitations of each~\cite{ft-vs-icl}.

\begin{table}[ht]
\centering
\begin{tabular}{@{}cccc@{}}
\toprule
Method & VQAv2 & OK-VQA & COCO \\ \midrule
LoRA & 55.60 & 47.06 & 97.75 \\
MimIC & 59.64 & 52.05 & 114.89 \\
MimIC + LoRA & \textbf{61.04} & \textbf{53.84} & \textbf{117.08} \\ \bottomrule
\end{tabular}
\caption{Performance of MimIC integrated with LoRA.}
\label{tab:mimic-lora}
\end{table}

\end{document}